\pdfoutput=1

\RequirePackage{snapshot}
\documentclass[11pt]{article}

\usepackage[]{ACL2023}

\usepackage{times}
\usepackage{latexsym}

\usepackage[T1]{fontenc}

\usepackage[utf8]{inputenc}

\usepackage{microtype}

\usepackage{inconsolata}

\usepackage{xspace}
\usepackage{graphicx}
\usepackage{nicefrac}
\usepackage{amsmath}
\usepackage{amssymb}
\usepackage{amsfonts}
\usepackage{amsthm}
\usepackage{mathtools}
\usepackage{booktabs}
\usepackage{tabularx}
\usepackage{multirow}
\usepackage{relsize}
\usepackage{enumitem}
\usepackage{makecell}
\usepackage{tikz-dependency}
\usepackage{textcomp}
\usepackage{microtype}
\usepackage{ifmtarg}
\usepackage{bussproofs}
\usepackage{arydshln}
\usepackage[linguistics]{forest}
\forestset{default preamble={
for tree={inner ysep = 0, outer sep = 2pt, font=\scriptsize,fit=tight,
}}}

\usepackage{url}

\usetikzlibrary{automata, fit, positioning, matrix, arrows.meta, decorations.pathmorphing, shapes.multipart}
\tikzset{
initial text=$ $,
}

\usepackage{algorithm}
\usepackage[noend]{algpseudocode}
\algrenewcommand\algorithmicindent{0.90em}%
\renewcommand\algorithmicdo{:}
\renewcommand\algorithmicthen{:}
\algnewcommand\algorithmicassert{\textbf{assert}}
\algnewcommand\Assert[1]{\State \algorithmicassert\ #1}
\newcommand{\rightcomment}[1]{{\color{gray} \(\triangleright\) {\footnotesize\textit{#1}}}}
\algrenewcommand{\algorithmiccomment}[1]{\hfill \rightcomment{#1}}
\algnewcommand{\LineComment}[1]{\State \rightcomment{#1}}
\algnewcommand{\LinesComment}[1]{\State \rightcomment{\parbox[t]{\linewidth-\leftmargin-\widthof{\(\triangleright\) }}{#1}}}
\algnewcommand{\IIf}[2]{%
  \State \algorithmicif\ #1\ \algorithmicthen\ #2}
\algnewcommand{\IIfElse}[3]{%
  \IIf{#1}{#2}\ \algorithmicelse\ #3}
\algnewcommand{\IElse}[1]{%
  \State \algorithmicelse\ #1}
\algnewcommand{\IFor}[2]{\State \algorithmicfor\ #1\ \algorithmicdo\ #2} 
\algnewcommand{\IWhile}[2]{\State \algorithmicwhile\ #1\ \algorithmicdo\ #2} 

\newcommand{\algorithmicfunc}[1]{\textbf{def} #1 :}
\algdef{SE}[FUNC]{Func}{EndFunc}[1]{\algorithmicfunc{#1}}{}
\makeatletter
\ifthenelse{\equal{\ALG@noend}{t}}%
  {\algtext*{EndFunc}}
  {}%
\makeatother

\newcommand{\checkNotation}[1]{{#1}}

\newcommand{\defn}[1]{\textbf{#1}}

\renewcommand{\hat}[1]{\widehat{#1}}
\renewcommand{\bar}[1]{\overline{#1}}
\newcommand{\tuple}[1]{\checkNotation{\langle #1 \rangle}}

\makeatletter
\newcommand*\bigcdot{\mathpalette\bigcdot@{.7}}
\newcommand*\bigcdot@[2]{\mathbin{\vcenter{\hbox{\scalebox{#2}{$\m@th#1\bullet$}}}}}
\makeatother

\newcommand{\semiring}{\checkNotation{\mathcal{S}}}
\newcommand{\Z}{\checkNotation{\mathrm{Z}}}

\newcommand{\mat}[1]{\checkNotation{\mathbf{#1}}}

\newcommand{\vx}{\mat{x}}
\newcommand{\vy}{\mat{y}}
\newcommand{\vz}{\mat{z}}
\newcommand{\x}[1]{{x}_{#1}}
\newcommand{\spanx}[1]{\checkNotation{\mat{x}_{#1}}}

\newcommand{\e}[1]{e_{#1}}

\newcommand{\defeq}[0]{\mathrel{\stackrel{\textnormal{\tiny def}}{=}}}

\newcommand{\bigo}[1]{\mathcal{O}\!\left(#1\right)}
\newcommand{\abs}[1]{\lvert #1 \rvert}

\newcommand{\nN}{\checkNotation{N}}

\newcommand{\nK}{\checkNotation{K}}

\newcommand{\equal}{{\,\textsf{=}\,}}

\newcommand{\algFace}[1]{\texttt{#1}\xspace}

\newcommand{\cfg}{\checkNotation{\mathcal{G}}}
\newcommand{\terminals}{\checkNotation{\Sigma}}
\newcommand{\nonterminals}{\checkNotation{\mathcal{N}}}
\newcommand{\rules}{\checkNotation{\mathcal{R}}}
\newcommand{\start}{\checkNotation{S}}
\newcommand{\tree}{\checkNotation{T}}
\newcommand{\trees}{\checkNotation{\mathcal{T}}}
\newcommand{\edge}[2]{\checkNotation{#1 \rightarrow #2}}

\newcommand{\dtree}[1]{\checkNotation{d}_{#1}}
\newcommand{\dtrees}[1]{\checkNotation{\mathcal{D}_{#1}}}

\newcommand{\subtrees}[1]{\trees^{#1}}

\newcommand{\emptystring}{\checkNotation{\varepsilon}}

\newcommand{\fsa}{\checkNotation{\mathcal{M}}}

\newcommand{\fsaEdges}{\checkNotation{\mathcal{E}}}

\newcommand{\dfaStates}{\checkNotation{\mathcal{Q}}}
\newcommand{\dfaAlpha}{\checkNotation{\mathcal{A}}}

\newcommand{\dfaStart}{\checkNotation{\mathcal{I}}}
\newcommand{\dfaFinish}{\checkNotation{\mathcal{F}}}

\newcommand{\dfaEdge}[3]{\checkNotation{#1\overset{#2}{\rightsquigarrow}#3}}
\newcommand{\dfaEnd}[1]{\checkNotation{\hat{#1}}}

\usepackage{tikz}
\usetikzlibrary{tikzmark}
\usetikzlibrary{calc}

\newcommand{\pgftextcircled}[1]{
    \setbox0=\hbox{#1}%
    \dimen0\wd0%
    \divide\dimen0 by 2%
    \begin{tikzpicture}[baseline=(a.base)]%
        \useasboundingbox (-\the\dimen0,0pt) rectangle (\the\dimen0,1pt);
        \node[circle,draw,outer sep=0pt,inner sep=0.1ex] (a) {#1};
    \end{tikzpicture}
}

\newcommand{\semiringtype}{\checkNotation{\mathbb{W}}}
\newcommand{\zero}[0]{\,{\smaller\pgftextcircled{\smaller $0$}}\,}
\newcommand{\one}[0]{\,{\smaller\pgftextcircled{\smaller $1$}}\,}

\newcommand{\semiringtuple}{\tuple{\semiringtype, \oplus, \otimes, \zero, \one}}

\newcommand{\stateset}[1]{\checkNotation{\mathcal{T}_{#1}}}
\newcommand{\predicted}[1]{\checkNotation{\mathcal{S}_{#1}}}

\newcommand{\chart}{\mathcal{C}}

\newcommand{\state}[1]{\checkNotation{[ #1 ]}}

\newcommand{\stategeneric}{\checkNotation{V}}
\newcommand{\stategenerictwo}{\checkNotation{U}}

\newcommand{\stdstate}{\state{i,j,\edge{A}{\mu \bigcdot \nu}}}

\newcommand{\stdstatecompleted}{\state{i,j,\edge{A}{\mu\, \nu\, \bigcdot}}}
\newcommand{\compstate}{\state{j, k, \edge{B}{\rho \,\bigcdot\,}}}
\newcommand{\compedstate}{\state{i, k, \edge{A}{\mu\, B \bigcdot \nu}}}
\newcommand{\scanstate}{\state{i, j, \edge{A}{\mu \bigcdot\, a\, \nu}}}
\newcommand{\scannedstate}{\state{i, k, \edge{A}{\mu\, a \bigcdot \nu}}}
\newcommand{\predstate}{\state{i, j, \edge{A}{\mu \bigcdot B\, \nu}}}
\newcommand{\predstateC}{\state{i, j, \edge{A}{\mu \bigcdot C\, \nu}}}
\newcommand{\prededstate}{\state{j, j, \edge{B}{\,\bigcdot\, \rho}}}

\newcommand{\fillstate}{\state{j, j, \edge{B}{\,\bigcdot\, \star}}}

\newcommand{\closestate}{\state{j, k, \edge{B}{ \star \,\bigcdot\,}}}

\newcommand{\scannedstatep}{\state{i, j, \edge{A}{\mu\, a \bigcdot \nu}}}
\newcommand{\prefixstate}{\state{j, j}}
\newcommand{\prefixstatek}{\state{k, k}}

\newcommand{\specialS}{S'}
\newcommand{\startstate}{\state{0,0,\edge{\specialS}{\,\bigcdot\, S}}}
\newcommand{\finalstate}{\state{0,\nN,\edge{\specialS}{S \,\bigcdot\,}}}

\newcommand{\rewrites}[2]{\checkNotation{#1 \!\overset{*}{\Rightarrow}\! #2}}
\newcommand{\leftcorner}[2]{\checkNotation{#1 \!\overset{*}{\Rightarrow}_L\! #2}}
\newcommand{\transitiveleftcorner}[2]{\checkNotation{#1 \!\overset{+}{\Rightarrow}_L\! #2}}
\newcommand{\notrewrites}[2]{\checkNotation{#1 \!\not\overset{*}{\Rightarrow}\! #2}}

\newcommand{\incinside}{\checkNotation{\dot{\inside}}}

\newcommand{\inside}{\checkNotation{\beta}}

\newcommand{\prefixoutside}{\checkNotation{\dot{\alpha}}}

\newcommand{\freeweight}[1]{\checkNotation{Z_{#1}}}

\newcommand{\lrrule}{\checkNotation{\edge{B}{B \, \rho}}}
\newcommand{\lrstate}{\checkNotation{\state{j, j, \edge{B}{\bigcdot \, B \, \rho}}}}

\newcommand{\pqueuecol}{\mathcal Q}
\newcommand{\pqueuespan}[1]{\mathcal P_{#1}}
\newcommand{\pqueueitem}[2]{\mathcal N_{#1#2}}
\newcommand{\pqueuespanstd}{\pqueuespan{k}}
\newcommand{\pqueueitemstd}{\pqueueitem{j}{k}}

\newcommand{\prefixtrue}{p}

\newcommand{\ruleFace}[1]{\textsc{#1}\xspace}
\newcommand{\pred}{\ruleFace{Pred}}
\newcommand{\predict}{\ruleFace{Predict}}
\newcommand{\scan}{\ruleFace{Scan}}
\newcommand{\comp}{\ruleFace{Comp}}
\newcommand{\complete}{\ruleFace{Complete}}

\newcommand{\scanone}{\ruleFace{Scan1}}
\newcommand{\scantwo}{\ruleFace{Scan2}}

\newcommand{\predone}{\ruleFace{Pred1}}
\newcommand{\predtwo}{\ruleFace{Pred2}}

\newcommand{\predonemod}{\ruleFace{Pred1lr}}

\newcommand{\startrule}{\ruleFace{Start}}

\newcommand{\compone}{\ruleFace{Comp1}}
\newcommand{\comptwo}{\ruleFace{Comp2}}

\newcommand{\componea}{\ruleFace{Comp1a}}
\newcommand{\componeb}{\ruleFace{Comp1b}}
\newcommand{\comptwoa}{\ruleFace{Comp2a}}
\newcommand{\comptwob}{\ruleFace{Comp2b}}

\newcommand{\pos}{\ruleFace{Pos}}

\newcommand{\filter}{\ruleFace{Filter}}
\newcommand{\eps}{\ruleFace{Epsilon}}

\newcommand{\example}{\ruleFace{Example}}

\newcommand{\earleyDeduction}{\algFace{Earley}}
\newcommand{\earleyDeductionFast}{\algFace{EarleyFast}}
\newcommand{\earleyFSA}{\algFace{EarleyFSA}}

\newcommand{\weight}[1]{\checkNotation{w\!\left(#1\right)}}
\newcommand{\weightprime}[1]{\checkNotation{w'\!\left(#1\right)}}

\newcommand{\outsidecolor}{violet}
\newcommand{\insidecolor}{cyan}
\newcommand{\incinsidecolor}{blue}

\newcommand{\unseencolor}{gray}

\newcommand{\nt}[1]{\textit{#1}}
\newcommand{\term}[1]{\textit{#1}}

\newcommand{\timvCut}[1]{}
\newcommand{\saveforcameraready}[1]{\textcolor{red}{#1}}
\renewcommand{\saveforcameraready}[1]{#1}
\newcommand{\removedforpresubmission}[1]{}
\renewcommand{\removedforpresubmission}[1]{\textcolor{purple}{#1}}

\usepackage[stable,hang,flushmargin]{footmisc}

\newcolumntype{C}{>{\centering\arraybackslash}X}

\theoremstyle{definition}

\usepackage{cleveref}
\crefname{section}{\S}{\S\S}
\Crefname{section}{\S}{\S\S}
\crefname{table}{Table}{Tables}
\crefname{figure}{Fig.}{Figs.}
\crefname{algorithm}{Alg.}{Algs.}
\crefname{line}{Line}{Lines}
\crefname{appendix}{App.}{Apps.}
\crefformat{section}{\S#2#1#3}
\crefname{thm}{Theorem}{Theorems}
\crefname{prop}{Proposition}{Propositions}
\crefname{defin}{Definition}{Definitions}
\crefname{lemma}{Lemma}{Lemmas}
\crefname{cor}{Corollary}{Corollaries}
\crefname{equation}{equation}{equations}
\crefname{inefficiency}{Inefficiency}{Inefficiencies}

\everypar{\looseness=-1}

\makeatletter
\newcommand{\toprulealg}{\hrule height.8pt depth0pt \kern2pt}
\newcommand{\midrulealg}{\kern2pt\hrule\kern2pt}
\newcommand{\bottomrulealg}{\kern2pt\hrule\relax}%
\newcommand{\algcaption}[2][]{%
  \refstepcounter{algorithm}%
  \@ifmtarg{#1}
    {\addcontentsline{loa}{figure}{\protect\numberline{\thealgorithm}{\ignorespaces #2}}}
    {\addcontentsline{loa}{figure}{\protect\numberline{\thealgorithm}{\ignorespaces #1}}}%
  \toprulealg
  \textbf{\fname@algorithm~\thealgorithm}\ #2\par
  \midrulealg
}
\makeatother

\usepackage[disable]{todonotes}
\makeatletter
\newcommand*\iftodonotes{\if@todonotes@disabled\expandafter\@secondoftwo\else\expandafter\@firstoftwo\fi}
\makeatother
\newcommand{\noindentaftertodo}{\iftodonotes{\noindent}{}}

\newcommand{\Fixme}[2][]{\noindentaftertodo}
\newcommand{\Notewho}[3][]{\noindentaftertodo}

\newcommand{\Ryan}[2][]{\noindentaftertodo}

\newcommand{\Ran}[2][]{\noindentaftertodo}

\newcommand{\Jason}[2][]{\noindentaftertodo}

\newcommand{\Andreas}[2][]{\noindentaftertodo}

\title{Efficient Semiring-Weighted Earley Parsing}
\newcommand{\ucambridge}{\normalfont \text{3}}
\newcommand{\ethz}{\text{\normalfont 1}}
\newcommand{\mpi}{\normalfont \text{2}}
\newcommand{\jhu}{\normalfont \text{4}}

\author{%
Andreas Opedal$^{\ethz}$\textsuperscript{,}$^{\mpi}$~\;~\;~Ran Zmigrod$^{\ucambridge}$~\;~\;~\textbf{Tim Vieira}$^{\jhu}$ \\ \textbf{Ryan Cotterell}$^{\ethz}$~\;~\;~\textbf{Jason Eisner}$^{\jhu}$ \\ $^{\ethz}$ETH Z{\"u}rich~\;~\;~\;~$^{\mpi}$Max Planck ETH Center for Learning Systems\\ $^{\ucambridge}$University of Cambridge~\;~\;~\;~$^{\jhu}$Johns Hopkins University\\
\texttt{\href{mailto:andreas.opedal@inf.ethz.ch}{andreas.opedal@inf.ethz.ch}}~\;~\texttt{\href{mailto:rz279@cam.ac.uk}{rz279@cam.ac.uk}}~\;~\texttt{\href{mailto:tim.f.vieira@gmail.com}{tim.f.vieira@gmail.com}} \\  
\texttt{\href{mailto:ryan.cotterell@inf.ethz.ch}{ryan.cotterell@inf.ethz.ch}}~\;~\texttt{\href{mailto:jason@cs.jhu.edu}{jason@cs.jhu.edu}} \\ 
}

\allowdisplaybreaks

\begin{document}
\maketitle

\begin{abstract}
This paper provides a reference description, in the form of a deduction system, of \citeauthor{earley70}'s \citeyearpar{earley70} context-free parsing algorithm with various speed-ups.  
Our presentation includes a known worst-case runtime improvement from \citeauthor{earley70}'s $\bigo{\nN^3\abs{\cfg}\abs{\rules}}$, which is unworkable for the large grammars that arise in natural language processing, to $\bigo{\nN^3\abs{\cfg}}$, which matches the runtime of CKY on a binarized version of the grammar $\cfg$.  Here $\nN$ is the length of the sentence, $\abs{\rules}$ is the number of productions in $\cfg$, and $\abs{\cfg}$ is the total length of those productions.
We also provide a version that achieves runtime of $\bigo{\nN^3 \abs{\fsa}}$ with $\abs{\fsa} \leq \abs{\cfg}$ when the grammar is represented compactly as a single finite-state automaton $\fsa$ (this is partly novel).
We carefully treat the generalization to semiring-weighted deduction, preprocessing the grammar like \citet{stolcke95} to eliminate deduction cycles, and further generalize \citeauthor{stolcke95}'s method to compute the weights of sentence prefixes.  We also provide implementation details for efficient execution, ensuring that on a preprocessed grammar, the semiring-weighted versions of our methods have the same asymptotic runtime and space requirements as the unweighted methods, including sub-cubic runtime on some grammars.
\newline
\newline
\vspace{1.5em} 
\hspace{.5em}\includegraphics[width=1.25em,height=1.25em]{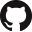}{\hspace{.75em}\parbox{\dimexpr\linewidth-2\fboxsep-2\fboxrule}{\url{https://github.com/rycolab/earleys-algo}}}
\end{abstract}

\section{Introduction}\label{sec:intro}

\Citet{earley70} was a landmark paper in computer science.\footnote{Based on the author's dissertation \cite{earley68}.}  Its algorithm was the first to directly parse under an \emph{unrestricted} context-free grammar in time $\bigo{\nN^3}$, with $\nN$ being the length of the input string.  Furthermore, it is faster for certain grammars because it uses left context to filter its search at each position. It parses unambiguous grammars in $\bigo{\nN^2}$ time and a class of ``bounded-state'' grammars, which includes all deterministic grammars, in $\bigo{\nN}$ time.  Its artful combination of top-down (goal-driven) and bottom-up (data-driven) inference later inspired a general method for executing logic programs, ``Earley deduction'' \cite{pereira-warren-1983-parsing}.

Earley's algorithm parses a sentence incrementally from left to right, optionally maintaining a packed parse forest over the sentence prefix that has been observed so far.  This supports online sentence processing---incremental computation of syntactic features and semantic interpretations---and also reveals for each prefix the set of grammatical choices for the next word.\footnote{In a programming language editor, incremental interpretation can support syntax checking, syntax highlighting, and tooltips; next-word prediction can support autocomplete.} 

It can be attractively extended to compute the \emph{probabilities} of the possible next words \cite{jelinek-lafferty-1991-computation, stolcke95}.  This is a standard way to compute autoregressive language model probabilities under a PCFG
to support cognitive modeling \citep{hale-2001-probabilistic} and speech recognition \citep{roark01}.
Such probabilities could further be combined with those of a large autoregressive language model to form a product-of-experts model.  Recent papers (as well as multiple github projects) have made use of a restricted version of this, restricting generation from the language model to only extend the current prefix in ways that are grammatical under an unweighted CFG; then only grammatical text or code will be generated \citep{semanticmachines-2021-emnlp, roy-2022-benchclamp, fang-2022-whole-truth}.

It is somewhat tricky to implement Earley's algorithm so that it runs as fast as possible.  Most importantly, the worst-case runtime should be linear in the size of the grammar, but this property was not achieved by  \citet{earley70} himself nor by textbook treatments of his algorithm \citep[e.g.,][\S13.4]{Jurafsky:2009:SLP:1214993}.  This is easy to overlook when the grammar is taken to be fixed, so that the grammar constant is absorbed into the $\mathcal{O}$ operator, as in the opening paragraph of this paper. 
Yet reducing the grammar constant is critical in practice, since natural language grammars can be very large \citep{dunlop2010reducing}.  For example, the Berkeley grammar \citep{petrov-etal-2006-learning}, a learned grammar for the Penn Treebank (PTB) \citep{marcusSM94}, contains over one million productions.\looseness=-1

In this reference paper, we attempt to collect the key efficiency tricks
and present them declaratively, in the form of a unified \emph{deduction system} that can be executed with good asymptotic complexity.\footnote{There has been no previous unified, formal treatment that is written as a deduction system, to the best of our knowledge. That said, declarative formulations have been presented in other formats in the dissertations of \citet{barthelemy-thesis}, \citet{clergerie-thesis}, and \citet{nederhof-thesis}.}
We obtain further speedups by allowing the grammar to be presented in the form of a weighted finite-state automaton whose paths correspond to the productions, which allows similar productions to share structure and thus to share computation.  Previous versions of this trick use a different automaton for each left-hand side nonterminal \citep[\textit{inter alia}]{purdom81,kochut1983towards,leermakers-1989-cover,perlin-1991-lr}; we show how to use a single automaton, which allows further sharing among productions with different left-hand sides.\looseness=-1

We carefully generalize our methods to handle semiring-weighted grammars, where the parser must compute the total weight of all trees that are consistent with an observed sentence \cite{goodman-1999-semiring}---or more generally, consistent with the prefix that has been observed so far.  Our goal is to ensure that if the semiring operations run in constant time, then semiring-weighted parsing runs in the same time and space as unweighted parsing (up to a constant factor), for \emph{every} grammar and sentence, including those where unweighted parsing is faster than the worst case.  \Citet{eisner-2023-tacl} shows how to achieve this guarantee for any acyclic deduction system, so we preprocess the grammar to eliminate cyclic derivations\footnote{Our method to remove nullary productions (\cref{app:null,app:fsa-null}) may be a contribution of this paper, as we were unable to find a correct construction in the literature.} (or nearly so: see \cref{app:twopass}).  Intuitively, this means we do not have to sum over infinitely many derivations at runtime (as \citet{goodman-1999-semiring} would).  
We also show how to compute prefix weights, which is surprisingly tricky and requires the semiring to be commutative.  Our presentation of preprocessing and prefix weights generalizes and corrects that of \citet{stolcke95}, who relied on special properties of PCFGs.

Finally, we provide a reference implementation in Cython\footnote{A fast implementation of Earley's algorithm is reported by \citet{polatSC16} but does not appear to be public.} and empirically demonstrate the value of the speedups.

\section{Weighted Context-Free Grammars}\label{sec:cfg}

\newcommand\Tstrut{\rule{0pt}{5ex}}
\newcommand\Bstrut{\rule[-3ex]{0pt}{0pt}}

\renewcommand{\arraystretch}{2.6}

\begin{table*}[t]
\centering
\small
\begin{tabularx}{0.98\linewidth}{l l l}
     & \earleyDeduction & \earleyDeductionFast  \\ \toprule 
     \vspace{-1pt}
    \bf Domains & \multicolumn{2}{l}{\hspace{0.6in}$i,j,k\in\{0,\dots,\nN\} \quad A,B\in\nonterminals \cup \{\specialS\} \quad a\in\terminals \quad \rho,\mu,\nu\in(\nonterminals\cup\terminals)^*$} \\
    \bf Items & \makecell[l]{$\stdstate \quad \state{j, k, a} \quad \edge{A}{\rho}$} &  \makecell[l]{$\stdstate \quad \state{j, k, a} \quad \edge{A}{\rho}$ \\
    $\state{i, j, \edge{A}{\,\bigcdot\, \star}}\hphantom{\mu\,} \quad \state{i, j, \edge{A}{\star \,\bigcdot\,}}$ \\
    } \vspace{5pt} \\
    \bf Axioms &  \makecell[l]{ $\edge{A}{\rho}, \forall (\edge{A}{\rho})\in\rules$ \\
    $\state{k-1, k, \x{k}}, \forall k \!\in\! \{1,\dots,\nN \}$ \\
    $\startstate$} &  \makecell[l]{$\edge{A}{\rho}, \forall (\edge{A}{\rho})\in\rules$ \\
    $\state{k-1, k, \x{k}}, \forall k \!\in\! \{1,\dots,\nN \}$ \\
    $\state{0,0,\edge{S}{\,\bigcdot\, \star}}$} \\
    \bf Goal &  $\finalstate$ & $\state{0,N,\edge{S}{\,\star\,\bigcdot}}$

    \\ \midrule
    &
        \AxiomC{$\edge{B}{\rho}$}
        \LeftLabel{\smaller \pred:}
        \RightLabel{$\predstate$}
        \UnaryInfC{$\prededstate$}
        \DisplayProof
    &    
        \AxiomC{}
        \RightLabel{$\predstate$}
        \LeftLabel{\smaller \predone:}
        \UnaryInfC{$\fillstate$}
        \DisplayProof
     \\
    &&        
        \AxiomC{$\edge{B}{\rho}$}
        \RightLabel{$\fillstate$}
        \LeftLabel{\smaller \predtwo:}
        \UnaryInfC{$\prededstate$}
        \DisplayProof \\
     \textbf{Rules} &
        \hspace{-7pt}
        \def\defaultHypSeparation{\hspace{0pt}}
        \AxiomC{$\scanstate$}
        \AxiomC{$\state{j, k, a}$}
        \LeftLabel{\smaller \scan:}
        \BinaryInfC{$\scannedstate$}
        \DisplayProof
    & 
        \hspace{-7pt}
        \def\defaultHypSeparation{\hspace{0pt}}
        \AxiomC{$\scanstate$}
        \AxiomC{$\state{j, k, a}$}
        \LeftLabel{\smaller \scan:}
        \BinaryInfC{$\scannedstate$}
        \DisplayProof
    \\
    &&
        \hspace{-7pt}
        \def\defaultHypSeparation{\hspace{0pt}}
        \AxiomC{$\compstate$}
        \LeftLabel{\smaller \compone:}
        \UnaryInfC{$\closestate$}
        \DisplayProof \\
    &
        \hspace{-7pt}
        \def\defaultHypSeparation{\hspace{0pt}}
        \AxiomC{$\predstate$}
        \AxiomC{$\compstate$}
        \LeftLabel{\smaller \comp:}
        \BinaryInfC{$\compedstate$}
        \DisplayProof
    & 
        \hspace{-7pt}
        \def\defaultHypSeparation{\hspace{0pt}}
        \AxiomC{$\predstate$}
        \AxiomC{$\closestate$}
        \LeftLabel{\smaller \comptwo:}
        \BinaryInfC{$\compedstate$}
        \DisplayProof 
    \\ \bottomrule
\end{tabularx}
\caption{Deduction systems for \citet{earley70}'s algorithm (\earleyDeduction) and our faster algorithm (\earleyDeductionFast).
An additional speedup is given in \cref{app:lookahead}.  Properly speaking, the ``items'' and ``rules'' shown here are templates; the actual items and rules are obtained by binding their variables to elements of their corresponding domains.}
\label{tab:systems}
\vspace{-1pt}
\end{table*}

A \defn{context-free grammar} (\defn{CFG}) $\cfg$ is a tuple $\tuple{\nonterminals, \terminals, \rules, \start}$ where $\terminals$ is a finite set of \defn{terminal} symbols,
$\nonterminals$ is a finite set of \defn{nonterminal} symbols with $\terminals \cap \nonterminals = \emptyset$, $\rules$ is a set of \defn{productions} from a nonterminal to a sequence of terminals and nonterminals (i.e., $\rules\subseteq\nonterminals\times(\terminals\cup\nonterminals)^*$), and $\start\in\nonterminals$ is the \defn{start} symbol.
We use lowercase variable names ($a, b, \dots$) and uppercase ones ($A, B, \dots$) for elements of $\terminals$  and $\nonterminals$, respectively.
We use a Greek letter ($\rho, \mu,$ or $\nu$) to denote a sequence of terminals and nonterminals, i.e., an element of $(\terminals\cup\nonterminals)^*$.
Therefore, a production has the form $\edge{A}{\rho}$.  Note that $\rho$ may be the empty sequence $\emptystring$.
We refer to $\abs{\rho} \geq 0$ as the \defn{arity} of the production, $\abs{\edge{A}{\rho}} \defeq 1+\abs{\rho}$ as the \defn{size} of the production, and $\abs{\cfg}\defeq\sum_{(\edge{A}{\rho})\in\rules}\abs{\edge{A}{\rho}}$
for the total \defn{size} of the CFG.
Therefore, if $\nK$ is the maximum arity of a production, $\abs{\cfg}\leq \abs{\rules}(1+\nK)$.
Productions of arity 0, 1, and 2 are referred to as \defn{nullary}, \defn{unary}, and \defn{binary} productions respectively.

For a given $\cfg$, we write $\mu \Rightarrow \nu$ to mean that $\mu \in (\terminals\cup\nonterminals)^*$ can be rewritten into $\nu \in (\terminals\cup\nonterminals)^*$ by a single production of $\cfg$. For example, $A\, B \Rightarrow \rho\, B$ expands $A$ into $\rho$ using the production $\edge{A}{\rho}$.  The reflexive and transitive closure of this relation, $\rewrites{\,}{\,}$, then denotes rewriting by any sequence of zero or more productions: for example, $\rewrites{A\, B}{\rho\, \mu\, \nu}$. We may additionally write $\leftcorner{A}{\rho}$ iff $\rewrites{A}{\rho \, \mu}$\saveforcameraready{, and refer to $\rho$ as a \defn{prefix} of $\rho \, \mu$}.

A \defn{derivation subtree} of $\cfg$ is a finite rooted ordered tree $\tree$ such that each node is labeled either with a terminal $a \in \terminals$, in which case it must be a leaf, or with a nonterminal $A \in \nonterminals$, in which case $\rules$ must contain the production $\edge{A}{\rho}$ where $\rho$ is the sequence of labels on the node's 0 or more children.
For any $A \in \nonterminals$, we write $\subtrees{A}$ for the set of derivation subtrees whose roots have label $A$, and refer to the elements of $\subtrees{S}$ as \defn{derivation trees}.
Given a string $\vx \in \terminals^*$ of length $\nN$, we write $\subtrees{A}_\vx$ for the set of derivation subtrees with leaf sequence $\vx$.  For an \defn{input sentence} $\vx$, its set of derivation trees $\trees_{\vx} \defeq \subtrees{S}_\vx$ is countable and possibly infinite.  It is non-empty iff $\rewrites{S}{\vx}$, with each $\tree \in \trees_{\vx}$ serving as a witness that $\rewrites{S}{\vx}$, i.e., that $\cfg$ can generate $\vx$.

We will also consider \defn{weighted CFG}s (\defn{WCFG}s), in which each production $\edge{A}{\rho}$ is additionally equipped with a \defn{weight} $\weight{\edge{A}{\rho}}\in\semiringtype$ where $\semiringtype$ is the set of values of a \defn{semiring}
$\semiring\defeq\semiringtuple$. Semirings are defined in \cref{app:semiring}.  We assume that $\otimes$ is commutative, deferring the trickier non-commutative case to \cref{app:noncomm}.
Any derivation tree $T$ of $\cfg$ can now be given a weight
\begin{equation}\label{eq:treeprod}
    \weight{\tree} \defeq \smashoperator{\bigotimes_{(\edge{A}{\rho})\in\tree}} \weight{\edge{A}{\rho}}
\end{equation}
where $\edge{A}{\rho}$ ranges over the productions associated with the nonterminal nodes of $T$.
The goal of a \defn{weighted recognizer} is to find the total weight of all derivation trees of a given input sentence $\vx$:
\begin{equation}\label{eq:z-input}
    \Z_{\vx} \defeq \weight{\rewrites{S}{\vx}} \defeq \smashoperator{\bigoplus_{\tree\in\trees_{\vx}}} \weight{\tree}
\end{equation}

\noindent
An ordinary \defn{unweighted recognizer} is the special case where $\semiringtype$ is the boolean semiring, so $\Z_{\vx} = \mathrm{true}$ iff $\rewrites{S}{\vx}$ iff $\trees_\vx \neq \emptyset$.
A \defn{parser} returns at least one derivation tree from $\trees_\vx$ iff $\trees_\vx\neq\emptyset$.

As an extension to the weighted recognition problem \labelcref{eq:z-input}, one may wish to find the \defn{prefix weight} of a string $\vy \in \terminals^*$, which is the total weight of all sentences $\vx=\vy\vz \in \terminals^*$ having that prefix:\looseness=-1
\begin{equation}\label{eq:prefix-weight}
    \weight{\leftcorner{S}{\vy}} \defeq \smashoperator{\bigoplus_{\vz\in\terminals^*}} \weight{\rewrites{S}{\vy \vz}} 
\end{equation}
\Cref{sec:intro} discussed applications of \defn{prefix probabilities}---the special case of \labelcref{eq:prefix-weight} for a \defn{probabilistic CFG} (\defn{PCFG}), in which the production weights are rewrite probabilities: $\semiringtype=\mathbb{R}_{\geq 0}$ and $(\forall A \in \nonterminals)
\sum_{(\edge{A}{\rho})\in\rules}\weight{\edge{A}{\rho}} = 1$.

\section{Parsing as Deduction}\label{sec:deduction}

We will describe Earley's algorithm using a \defn{deduction system}, 
a formalism that is often employed in the presentation of parsing algorithms \citep{pereira1987prolog,sikkel97}, as well as in mathematical logic and programming language theory \citep{pierce}.  Much is known about how to execute \cite{goodman-1999-semiring}, transform \cite{eisner-blatz-2007}, and neuralize \cite{mei-et-al-2020-icml} deduction systems.

A deduction system proves \defn{items} $\stategeneric$ using \defn{deduction rules}.  Items represent propositions; the rules are used to prove all propositions that are true.  
A deduction rule is of the form
\begin{prooftree}
\AxiomC{$U_1$}
\AxiomC{$U_2$}
\AxiomC{$\cdots$}
\LeftLabel{\small \example:}
\TrinaryInfC{$\stategeneric$}
\end{prooftree}
where \example is the name of the rule, the 0 or more items above the bar are called \defn{antecedents}, and the single item below the bar is called a \defn{consequent}.
Antecedents may also be written to the side of the bar; these are called \defn{side conditions} and will be handled differently for weighted deduction in \cref{sec:weighted}.
\defn{Axioms} (listed separately) are merely rules that have no antecedents; as a shorthand, we omit the bar in this case and simply write the consequent.

A \defn{proof tree} is a finite rooted ordered tree whose nodes are labeled with items, and where every node is licensed by the existence of a deduction rule whose consequent $\stategeneric$ matches the label of the node and whose antecedents $U_1,U_2,\ldots$ match the labels of the node's children.  It follows that the leaves are labeled with axioms.  A \defn{proof} of item $\stategeneric$ is a proof tree $\dtree{\stategeneric}$ whose root is labeled with $\stategeneric$: this shows how $\stategeneric$ can be deduced from its children, which can be deduced from their children, and so on until axioms are encountered at the leaves.  We say $\stategeneric$ is \defn{provable} if $\dtrees{\stategeneric}$, which denotes the set of all its proofs, is nonempty.

Our unweighted recognizer determines whether a certain \defn{goal item} is provable by a certain set of deduction rules from axioms that encode $\cfg$ and $\vx$.  The deduction system is set up so that this is the case iff $\rewrites{S}{\vx}$.  The recognizer can employ a \defn{forward chaining} method \cite[see e.g.][]{ceri-et-al-1990,eisner-2023-tacl} that iteratively deduces items by applying deduction rules whenever possible to antecedent items that have already been proved; this will eventually deduce all provable items.  An unweighted parser extends the recognizer with some extra bookkeeping that lets it return one or more actual proofs of the goal item if it is provable.\footnote{Each proved item stores a ``backpointer'' to the rule that proved it.  Equivalently, an item's proofs may be tracked by its weight in a ``derivation semiring'' \citep{goodman-1999-semiring}.}

\section{Earley's Algorithm}\label{sec:earley}

Earley's algorithm can be presented as the specific deduction system \earleyDeduction shown in \cref{tab:systems} \cite{sikkel97,shieber-schabes-pereira-1995,goodman-1999-semiring}, explained in more detail in \cref{app:earley-original}.  Its proof trees $\dtrees{\text{goal}}$ are in one-to-one correspondence with the derivation trees $\trees_\vx$ (a property that we will maintain for our improved deduction systems in \cref{sec:new} and \cref{sec:fsa}).
The grammar $\cfg$ is encoded by axioms $\rewrites{A}{\rho}$ that correspond to the productions of the grammar.
The input sentence $\vx$ is encoded by axioms of the form $\state{k-1, k, a}$ where $a\in\terminals$; this axiom is true iff $\spanx{k-1:k} = x_{k} = a$.\footnote{\label{fn:lattice}All methods in this paper can be also applied directly to lattice parsing, in which $i,j,k$ range over states in an acyclic lattice of possible input strings, and $0$ and $N$ refer to the unique initial and final states.  A lattice edge from $j$ to $k$ labeled with terminal $a$ is then encoded by the axiom $\state{j, k, a}$.}
The remaining items have the form $\stdstate$, where 
$0 \leq i \leq j \leq N$, so that the \defn{span} $(i, j)$ refers to a substring $\spanx{i:j} \defeq x_{i+1}\cdots x_j$ of the input sentence $\vx=x_1 x_2 \ldots x_N$.
The item $\stdstate$ is derivable only if the grammar $\cfg$ has a production $\edge{A}{\mu\, \nu}$ such that $\rewrites{\mu}{\spanx{i:j}}$.
Therefore, $\bigcdot$ indicates the progress we have made through the production.
An item with nothing to the right of $\bigcdot$, e.g., $\state{i,j,\edge{A}{\rho\,\bigcdot\,}}$, is called \defn{complete}.
The set of all items with a shared right index $j$ is called the \defn{item set} of $j$, denoted $\stateset{j}$.\looseness=-1

While $\rewrites{\mu}{\spanx{i:j}}$ is a necessary condition for $\stdstate$ to be provable, it is not sufficient.  For efficiency, the \earleyDeduction deduction system is cleverly constructed so that this item is provable iff\footnote{\label{fn:coacc}Assuming that all nonterminals $B\in\nonterminals$ are \defn{generating}, i.e., $\exists\vx' \in \terminals^*$ such that $\rewrites{B}{\vx'}$.  To ensure this, repeatedly mark $B\in\nonterminals$ as generating whenever $\rules$ contains some $\edge{B}{\rho}$ such that all nonterminals in $\rho$ are already marked as generating.  Then delete any unmarked nonterminals and their rules.\looseness=-1}
it can appear in a proof of the goal item for some input string beginning with $\spanx{0:j}$, and thus possibly for $\vx$ itself.\saveforcameraready{\footnote{\citet{earley70} also generalized the algorithm to prove this item only if it can appear in a proof of some string that begins with $\spanx{0:(j+\Delta)}$, for a fixed $\Delta$.  This is lookahead of $\Delta$ tokens.}}\looseness=-1

Including $\startstate$ as an axiom in the system effectively causes forward chaining to start looking for a derivation at position $0$.   Forward chaining will prove the goal item $\finalstate$ 
iff $\rewrites{S}{\vx}$.  These two items conveniently pretend that the grammar has been augmented with a new start symbol $\specialS \notin \nonterminals$ that only rewrites according to the single production $\edge{\specialS}{S}$.  

The \earleyDeduction system employs three deduction rules: \predict, \scan, and \complete.
We refer the reader to \cref{app:earley-original} for a presentation and analysis of these rules, which reveals a total runtime of $\bigo{\nN^3\abs{\cfg}\abs{\rules}}$.
\Cref{sec:past} outlines how past work improved this runtime. In particular, \citet{grahamHR80} presented an unweighted recognizer that is a variant of Earley's, along with implementation details that enable it to run in time $\bigo{\nN^3\abs{\cfg}}$. However, those details were lost in retelling their algorithm as a deduction system \cite[p.~113]{sikkel97}.
Our improved deduction system in the next section does enable the  $\bigo{\nN^3\abs{\cfg}}$ runtime, with execution details of forward chaining spelled out in \cref{sec:implementation}.

\section{An Improved Deduction System}\label{sec:new}

Our \earleyDeductionFast deduction system, shown in the right column of \cref{tab:systems}, shaves a factor of $\bigo{\rules}$ from the runtime of \earleyDeduction. It does so by effectively applying a weighted fold transform \citep{fold, eisner-blatz-2007, johnson-2007-transforming} on \pred (\cref{sec:new-pred}) and \comp (\cref{sec:new-comp}), introducing
coarse-grained items of the forms $\state{i, j, \edge{A}{\,\bigcdot\, \star}}$ and $\state{i, j, \edge{A}{\star \,\bigcdot\,}}$.  
In these items, the constant symbol $\star$ can be regarded as 
a wildcard that stands for ``any sequence $\rho$.''
We also use these new items to replace the goal item and the axiom that used $\specialS$; the extra $\specialS$ symbol is no longer needed.
The proofs are essentially unchanged (\cref{app:one-to-one}).

We now describe our new deduction rules for \comp and \pred.  (\scan is unchanged.)  We also analyze their runtime, using the same techniques as in \cref{app:earley-original}.

\subsection{Predict}\label{sec:new-pred}

We split \pred into two rules: \predone and \predtwo.
The first rule, \predone, creates an item that gathers together all requests to look for a given nonterminal $B$ starting at a given position $j$:

\begin{prooftree}
\AxiomC{}
\RightLabel{$\predstate$}
\LeftLabel{\smaller \predone:}
\UnaryInfC{$\fillstate$}
\end{prooftree}
There are three free choices in the rule: indices $i$ and $j$, and dotted production $\edge{A}{\,\bigcdot\, B\, \nu}$.
Therefore, \predone has a total runtime of $\bigo{\nN^2\abs{\cfg}}$.

The second rule, \predtwo,
expands the item into commitments to look for each specific kind of $B$:
\begin{prooftree}
\AxiomC{$\edge{B}{\rho}$}
\LeftLabel{\smaller \predtwo:}
\RightLabel{$\fillstate$}
\UnaryInfC{$\prededstate$}
\end{prooftree}
\predtwo has two free choices: index $j$ and production $\edge{B}{\rho}$.
Therefore, \predtwo has a runtime of $\bigo{\nN\abs{\rules}}$, which is dominated by $\bigo{\nN\abs{\cfg}}$ and so the two rules together have a runtime of $\bigo{\nN^2\abs{\cfg}}$.

\subsection{Complete}\label{sec:new-comp}
We speed up \comp in a similar fashion to \pred.
We split \comp into two rules: \compone and \comptwo.
The first rule, \compone, 
gathers all complete $B$ constituents over a given span into a single item:  
\begin{prooftree}
\AxiomC{$\compstate$}
\LeftLabel{\smaller \compone:}
\UnaryInfC{$\closestate$}
\end{prooftree}
We have three free choices: indices $j$ and $k$, and complete production $\edge{B}{\rho}$ with domain size $\abs{\rules}$.
Therefore, \compone has a total runtime of $\bigo{\nN^2\abs{\rules}}$, or $\bigo{\nN^2\abs{\cfg}}$.

The second rule, \comptwo, attaches the resulting complete items to any incomplete items that predicted them:
\begin{prooftree}
\def\defaultHypSeparation{\hspace{0pt}}
\AxiomC{$\predstate$}
\AxiomC{$\closestate$}
\LeftLabel{\smaller \comptwo:}
\BinaryInfC{$\compedstate$}
\end{prooftree}
We have four free choices: indices $i$, $j$, and $k$, and dotted production $\edge{A}{\mu \bigcdot B\, \nu}$.
Therefore, \comptwo has a total runtime of $\bigo{\nN^3\abs{\cfg}}$ and so the two rules together have a runtime of $\bigo{\nN^3\abs{\cfg}}$.

\section{Semiring-Weighted Parsing}\label{sec:weighted}
We have so far presented Earley's algorithm and our improved deduction system in the unweighted case.
However, we are often interested in determining not just whether a parse exists, but the total weight of all parses as in \cref{eq:z-input}, or the total weight of all parses consistent with a given prefix as in \cref{eq:prefix-weight}.

We first observe that by design, the derivation trees of the CFG are in 1-1 correspondence with the proof trees of our deduction system that are rooted at the goal item. Furthermore, the weight of a derivation subtree can be found as the weight of the corresponding proof tree, if the weight $\weight{\dtree{\stategeneric}}$ of any proof tree $\dtree{\stategeneric}$ is defined recursively as follows.

\textbf{Base case:} $\dtree{\stategeneric}$ may be a single node, i.e., $\stategeneric$ is an axiom.  If $\stategeneric$ has the form $\edge{A}{\rho}$, then $\weight{\dtree{\stategeneric}}$ is the weight of the corresponding grammar production, i.e., $\weight{\edge{A}{\rho}}$. All other axiomatic proof trees of \earleyDeduction and \earleyDeductionFast have weight $\one$.\footnote{However, this will not be true in \earleyFSA (\cref{sec:fsa} below).  There the grammar is given by a WFSA, and each axiom corresponding to an arc or final state of this grammar will inherit its weight from that arc or final state.  Similarly, if we generalize to lattice parsing---where the \emph{input} is given by an acyclic WFSA and each proof tree corresponds to a parse of some weighted path from this so-called lattice---then an axiom providing a terminal token should use the weight of the corresponding lattice edge.  Then the weight of the proof tree will include the total weight of the lattice path along with the weight of the CFG productions used in the parse.}

\textbf{Recursive case:} If the root node of $\dtree{\stategeneric}$  
has child subtrees $\dtree{U_1}, \dtree{U_2}, \ldots$, then $\weight{\dtree{\stategeneric}} = \weight{\dtree{U_1}}\otimes\weight{\dtree{U_2}}\otimes\cdots$.  However, the factors in this product include only the antecedents written above the bar, not the side conditions (see \cref{sec:deduction}).  

Following \citet{goodman-1999-semiring}, we may also 
associate a weight with each item $\stategeneric$, denoted $\incinside(\stategeneric)$, which is the \emph{total} weight of \emph{all} its proofs $\dtree{\stategeneric} \in \dtrees{\stategeneric}$. By the distributive property, we can obtain that weight as an $\oplus$-sum over all one-step proofs of $\stategeneric$ from antecedents.  Specifically, each deduction rule that deduces $\stategeneric$ contributes an $\oplus$-summand, given by the product $\incinside(U_1)\otimes\incinside(U_2)\otimes\cdots$ of the weights of its antecedent items (other than side conditions).  

Now our weighted recognizer can obtain $Z_{\vx}$ (the total weight of all derivations of $\vx$) as $\incinside$ of the goal item (the total weight of all proofs of that item).

For an item $\stategeneric$ of the form $\stdstate$, the weight $\incinside(\stategeneric)$ will consider derivations of nonterminals in $\mu$ but not those in $\nu$. We therefore refer to $\incinside(\stategeneric)$ as an \defn{incomplete inside weight}.  However,
$\nu$ will come into play in the extension of \cref{sec:prefix}.

The deduction systems 
work for any semiring-weighted CFG.
Unfortunately, the forward-chaining algorithm for \emph{weighted} deduction 
\cite[Fig.\@ 3]{eisner-etal-2005-compiling}
may not terminate if the system permits \emph{cyclic} proofs, where an item can participate in one of its own proofs.
In this case, the algorithm will merely approach the correct value of $Z_\vx$ as it discovers deeper and deeper proofs of the goal item. 
Cyclicity in our system can arise from sets of unary productions such as $\{\edge{A}{B}, \edge{B}{A}\} \subseteq \rules$, or equivalently, from $\{\edge{A}{E\,B\,E}, \edge{B}{A}\} \subseteq \rules$ where $\rewrites{E}{\emptystring}$ (which is possible if $\rules$ contains $\edge{E}{\emptystring}$ or other nullary productions).
We take the approach of eliminating problematic unary and nullary productions from the weighted grammar without changing $Z_\vx$ for any $\vx$.
We provide methods to do this in \cref{app:unary} and \cref{app:null} respectively.
It is important to eliminate nullary productions \emph{before} eliminating unary cycles, since nullary removal may create new unary productions. 
The elimination of some productions can increase $\abs{\cfg}$, but we explain how to limit this effect.

\subsection{Extension to Prefix Weights}
\label{sec:prefix}

\citet{stolcke95} showed how to extend Earley's algorithm to compute prefix probabilities under PCFGs, by associating a ``forward probability'' with each $\bigcdot$-item.\footnote{
Also other CFG parsing algorithms can be adapted to compute prefix probabilities, e.g., CKY 
\citep{jelinek-lafferty-1991-computation, nowak-cotterell-2023}. }
However, he relied on the property that all nonterminals $A$ have $\freeweight{A}=1$, where $\freeweight{A}$ denotes the \defn{free weight} 
\begin{align}\label{eq:free-weight}
    \freeweight{A} &\defeq \bigoplus_{\tree \in \subtrees{A}} \bigotimes_{\edge{B}{\rho} \in \tree} \weight{\edge{B}{\rho}}
\end{align}
As a result, his algorithm does not handle the case of WCFGs
or CRF-CFGs \cite{johnson-etal-1999-estimators, yusuke-02-maximum, finkel-kleeman-manning-2008}, or even non-tight
PCFGs \cite{chi-geman-1998}.  It also does not handle
semiring-weighted grammars.  We generalize by associating with each
$\bigcdot$-item, instead of a ``forward probability,'' a ``prefix outside
  weight'' from the same commutative semiring that is used to weight the grammar productions.  Formally, each $\weight{\stategeneric}$ will now be a pair $(\incinside(\stategeneric), \prefixoutside(\stategeneric))$, and we combine these pairs in specific ways.

\begin{figure}[t!]

\centering
     \includegraphics[width=1.00\linewidth]{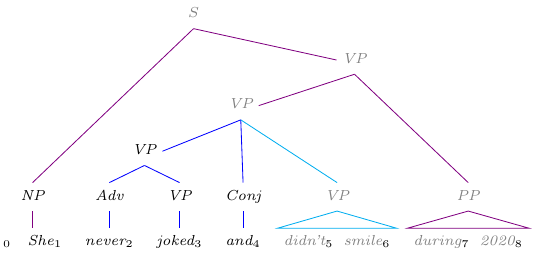}
\caption{
This derivation tree corresponds to a proof under \earleyDeductionFast.  Since $\state{1, 4, \edge{\nt{VP}}{\nt{VP}\ \nt{Conj} \bigcdot \nt{VP}}}$ is a $\bigcdot$-item in that proof, it partitions the steps of that proof into \textcolor{\incinsidecolor}{incomplete inside}, \textcolor{\insidecolor}{future inside} and \textcolor{\outsidecolor}{outside} portions that respectively prove $\state{1, 4, \edge{\nt{VP}}{\nt{VP}\ \nt{Conj} \bigcdot \nt{VP}}}$, prove the complete item $\state{1, 6, \edge{\nt{VP}}{\nt{VP}\ \nt{Conj}\ \nt{VP} \bigcdot \mbox{}}}$ from that, and prove goal from the complete item.  See \cref{fig:prefix-tree} for an alternative derivation and more discussion.
}\label{fig:prefix-tree-main}
\end{figure}

Recall from \cref{sec:earley} that the item $\stategeneric = \stdstate$ is provable iff\textsuperscript{\ref{fn:coacc}} it appears in a proof of some sentence beginning with $\spanx{0:j}$.  For any such proof containing $\stategeneric$, its steps can be partitioned as shown in \cref{fig:prefix-tree-main}, factoring the proof weight into three factors.  Just as the incomplete inside weight $\incinside(\stategeneric)$ is the total weight of all ways to prove $\stategeneric$, the \defn{future inside weight} $Z_\nu$ is the total weight of all ways to prove $\stdstatecompleted$ from $V$and the \defn{prefix outside weight} $\prefixoutside(\stategeneric)$ is the total weight of all ways to prove the goal item from $\stdstatecompleted$---in both cases allowing \emph{any} future words $\spanx{j:}$ as ``free'' axioms.\footnote{Prefix outside weights differ from traditional outside weights \citep{baker-1979,lari-1990-estimation, eisner16backprop}, which restrict to the \emph{actual} future words $\spanx{j:n}$.
} 

The future inside weight $Z_\nu=\prod_{i: \nu_i \in \nonterminals} Z_{\nu_i}$ does not depend on the input sentence.  To avoid a slowdown at parsing time, we precompute this product for each suffix $\nu$ of each production in $\rules$, after using methods in \cref{app:null} to precompute the free weights $\freeweight{A}$ for each nonterminal $A$. 

Like $\incinside(\stategeneric)$, $\prefixoutside(\stategeneric)$ is obtained as an $\oplus$-sum over all one-step proofs of $\stategeneric$.  Typically, each one-step proof increments $\prefixoutside(\stategeneric)$ by the prefix outside weight of its $\bigcdot$-antecedent or $\bigcdot$-side condition (for $\comptwo$, the \emph{left} $\bigcdot$-antecedent).  As an important exception, when $V = \fillstate$, each of its one-step proofs via \predone instead increments $\prefixoutside(\stategeneric)$ by
\begin{align}
  & \prefixoutside(\predstate)
  \nonumber \\
  & \mbox{} \otimes \incinside(\predstate) \otimes Z_\nu \label{eq:narrow-outside}
\end{align}
combining the steps outside $\predstate$ with some steps inside the $A$ (including its production) to get all the steps outside the $B$.  The base case is the start axiom, $\prefixoutside(\state{0,0,\edge{S}{\,\bigcdot\, \star}}) = \one$.

Unfortunately, this computation of $\prefixoutside(\stategeneric)$ is only correct if there is no left-recursion in the grammar.  We explain this issue in \cref{app:left-recursion} and fix it by extending the solution of \citet[\S4.5.1]{stolcke95}.

The prefix weight of $\spanx{0:j}$ $(j > 0)$ is computed as an $\oplus$-sum
$\prefixoutside(\prefixstate)$ over all one-step proofs of the new item $\prefixstate$ via the following new deduction rule that is triggered by the consequent of \scan:
\begin{prooftree}
\AxiomC{}
\LeftLabel{\smaller \pos:}
\RightLabel{$\scannedstatep$}
\UnaryInfC{$\prefixstate$}
\end{prooftree} 
Each such proof increments the prefix weight
by 
\begin{align}
  & \prefixoutside(\scannedstatep)
  \nonumber \\
  & \mbox{} \otimes \incinside(\scannedstatep) \otimes Z_\nu \label{eq:pos-rule}
\end{align}

\section{Earley's Algorithm Using an FSA}\label{sec:fsa}
\renewcommand{\arraystretch}{2.4}

\begin{table*}[t]
\centering
\small
\begin{tabularx}{0.98\linewidth}{l l l}
    \toprule
    \bf  Domains & \multicolumn{2}{l}{$i,j,k\in\{0,\dots,\nN\} \quad A\in\nonterminals \quad a\in\terminals \quad q,q'\in\dfaStates$} \\
    \bf Items & \multicolumn{2}{l}{$\state{i, j, q} \; \state{i, j, q?} \; \state{i, j, a} \; \state{i, j, \edge{A}{\,\bigcdot\, \star}} \; \state{i, j, \edge{A}{\star \,\bigcdot\,}} \; \underbrace{q\in\dfaStart \quad q'\in\dfaFinish \quad \dfaEdge{q}{a}{q'} \quad \dfaEdge{q}{A}{q'} \quad \dfaEdge{q}{A}{\star} \quad \dfaEdge{q}{*\hat{A}}{\star}}_{\text{WFSA items}}$} \\ 
    \bf Axioms &  \multicolumn{2}{l}{$\state{k-1, k, \x{k}}, \forall k \!\in\! \{1,\dots,\nN \} \quad\quad \state{0, 0, \edge{S}{\bigcdot\,\star}}$\quad WFSA items derived from the WFSA grammar (see \cref{sec:fsa})} \\ 
    \bf Goals & $\state{0, \nN, \edge{S}{\star \,\bigcdot\,}}$

    \\ \midrule
    & 
        \AxiomC{}
        \RightLabel{\makecell[l]{$\state{i, j, q}$ \\ $\dfaEdge{q}{B}{\star}$}}
        \LeftLabel{\smaller \predone:}
        \UnaryInfC{$\state{j, j, \edge{B}{\,\bigcdot\, \star}}$}
        \DisplayProof
    & 
        \def\defaultHypSeparation{\hspace{0pt}}
        \AxiomC{$\state{j, k, q}$}
        \AxiomC{$\dfaEdge{q}{\dfaEnd{B}}{q'}$}
        \AxiomC{$q'\in\dfaFinish$}
        \LeftLabel{\smaller \compone:}
        \TrinaryInfC{$\state{j, k, \edge{B}{ \star \,\bigcdot\,}}$}
        \DisplayProof \\
    \textbf{Rules}        
    & 

        \AxiomC{$q\in\dfaStart$}
        \LeftLabel{\smaller \predtwo:}
        \UnaryInfC{$\state{j, j, q?}$}
        \DisplayProof
    & 
        \def\defaultHypSeparation{\hspace{0pt}}
        \AxiomC{$\state{i, j, q}$}
        \AxiomC{$\dfaEdge{q}{B}{q'}$}
        \AxiomC{$\state{j, k, \edge{B}{ \star \,\bigcdot\,}}$}
        \LeftLabel{\smaller \comptwo:}
        \TrinaryInfC{$\state{i, k, q'?}$}
        \DisplayProof

    \\
    & 
        \hspace{-7pt}
        \def\defaultHypSeparation{\hspace{0pt}}
        \AxiomC{$\state{i, j, q}$}
        \AxiomC{$\dfaEdge{q}{a}{q'}$}
        \AxiomC{$\state{j, k, a}$}
        \LeftLabel{\smaller \scan:}
        \TrinaryInfC{$\state{i, k, q'?}$}

        \DisplayProof
    &
        \hspace{-7pt}
        \def\defaultHypSeparation{\hspace{0pt}}
        \AxiomC{$\state{i, j, q}$}
        \AxiomC{$\dfaEdge{q}{\emptystring}{q'}$}
        \LeftLabel{\smaller \eps:}
        \BinaryInfC{$\state{i, j, q'?}$}
        \DisplayProof

        \def\defaultHypSeparation{\hspace{0pt}}
        \AxiomC{$\state{j, k, q?}$}
        \RightLabel{\makecell[l]{
        $\state{j, j, \edge{A}{\,\bigcdot\, \star}}$ \\ $\dfaEdge{q}{*\hat{A}}{\star}$}
        }
        \LeftLabel{\smaller \filter:}
        \UnaryInfC{$\state{j, k, q}$}
        \DisplayProof
    \\ \bottomrule
\end{tabularx}
\caption{\earleyFSA, a variant of \earleyDeductionFast in which FSA states replace dotted productions. Side conditions are stacked horizontally in the interest of space. A faster binarized version is given in supplementary material as \cref{tab:fsa-binarized}.\looseness=-1}
\label{tab:fsa}
\end{table*}

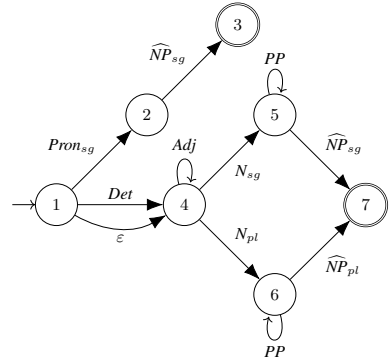
\begin{figure}[t] 
\centering
    \centering
    \small
    \scalebox{0.7}{
    \begin{tikzpicture}[node distance = 16mm]
    \node[state, initial] (q1) { $1$ }; 
    \node[state] (q2) [above right = of q1] { $2$ };
    \node[state, accepting] (q3) [above right = of q2] { $3$ }; 
    \node[state] (q4) [right = of q1] { $4$ };
    \node[state] (q5) [above right = of q4] { $5$ };
    \node[state] (q6) [below right = of q4] { $6$ };
    \node[state, accepting] (q7) [below right = of q5] { $7$ };
    \draw[-{Latex[length=3mm]}] 

    (q1) edge[left] node[above left]{ $\nt{Pron}_{sg}$ } (q2)
    (q2) edge[left] node[above left]{ $\hat{\nt{NP}}_{sg}$ } (q3)

    (q1) edge[right] node[above]{ $\nt{Det}$ } (q4) 
    (q1) edge[bend right] node[below]{ $\emptystring$ } (q4)
    (q4) edge[loop above] node[above]{ $\nt{Adj}$ } (q4)

    (q4) edge[left] node[below right]{ $\nt{N}_{sg}$ } (q5)
    (q5) edge[loop above] node[above]{ $\nt{PP}$ } (q5)
    (q5) edge[left] node[above right]{ $\hat{\nt{NP}}_{sg}$ } (q7)

    (q4) edge[right] node[above right]{ $\nt{N}_{pl}$ } (q6)
    (q6) edge[loop below] node[below]{ $\nt{PP}$ } (q6)
    (q6) edge[right] node[below right]{ $\hat{\nt{NP}}_{pl}$ } (q7)

    ;
    \end{tikzpicture}
    }
\vspace{-8pt}
\caption{Part of an FSA for English, showing ways to generate singular (sg) and plural (pl) noun phrases. The paths shown here correspond to the extended CFG production rules $\edge{\nt{NP}_{sg}}{(\ \nt{Det}?\ \nt{Adj}^*\ \nt{N}_{sg}^+\ \nt{PP}^*\ ) \mid \nt{Pron}_{sg}}$ and $\edge{\nt{NP}_{pl}}{\nt{Det}?\ \nt{Adj}^*\ \nt{N}_{pl}^+\ \nt{PP}^*}$.}\label{fig:earley-fsa-noun-phrases}
\end{figure}

In this section, we present a generalization of \earleyDeductionFast that can parse with any \defn{weighted finite-state automaton} (\defn{WFSA}) grammar $\fsa$ in $\bigo{\nN^3\abs{\fsa}}$. Here $\fsa$ is a WFSA \citep{Mohri2009}\saveforcameraready{} that encodes the CFG productions as follows.  For any $\rho \in (\terminals \cup \nonterminals)^*$ and any $A \in \nonterminals$, for $\fsa$ to accept the string $\rho\, \dfaEnd{A}$ with weight $w \in \semiringtype$ is tantamount to having the production $\edge{A}{\rho}$ in the CFG with weight $w$.  The grammar size $\abs{\fsa}$ is the number of WFSA arcs.  See \cref{fig:earley-fsa-noun-phrases} for an example.

This presentation has three advantages over a CFG.  First, $\fsa$ can be compiled from an extended CFG \citep{purdom81}, which allows user-friendly specifications like $\edge{\nt{NP}}{\nt{Det}?\ \nt{Adj}^*\ \nt{N}^+\ \nt{PP}^*}$ that may specify infinitely many productions with unboundedly long right-hand-sides $\rho$ (although $\fsa$ still only describes a context-free language).\
Second, productions with similar right-hand-sides can be partially merged to achieve a smaller grammar and a faster runtime.  They may share partial paths in $\fsa$, which means that a single item can efficiently represent many dotted productions.
Third, when $\otimes$ is non-commutative,\saveforcameraready{} only the WFSA grammar formalism allows elimination of nullary rules in all cases (see \cref{app:null}).\looseness=-1

Our WFSA grammar is similar to a \defn{recursive transition network} or \defn{RTN} grammar \citep{woods70}.
Adapting Earley's algorithm to RTNs was discussed by \citet{kochut1983towards}, \citet{leermakers-1989-cover}, and \citet{perlin-1991-lr}. \Citet{klein2001time} used a weighted version for PTB parsing. None of them spelled out a deduction system, however.

Also, an RTN is a collection of productions of the form $\edge{A}{\fsa_A}$, where for $\fsa_A$ to accept $\rho$ corresponds to having $\edge{A}{\rho}$ in the CFG.  Thus an RTN uses one FSA per nonterminal. Our innovation is to use one WFSA for the entire grammar, specifying the left-hand-side nonterminal as a final symbol.  Thus, to allow productions $\edge{A}{\mu\,\nu}$ and $\edge{B}{\mu\,\nu'}$, our single WFSA can have paths $\mu\,\nu\,\hat{A}$ and $\mu\,\nu'\,\hat{B}$ that share the $\mu$ prefix---as in \cref{fig:earley-fsa-noun-phrases}.  This allows our \earleyFSA to match the $\mu$ prefix only once, in a way that could eventually result in completing either an $A$ or a $B$ (or both).\footnote{\Citet{nederhof-1994-optimal} also shares prefixes between $A$ and $B$; but there, once paths split to yield separate items, they cannot re-merge to share a \emph{suffix}.  We can merge by deriving $\state{j,k,q?}$ in multiple ways. Our $\state{j,k,q?}$ does not specify its set of target left-hand sides; \filter recomputes that set dynamically.\looseness=-1}

A traditional weighted CFG $\cfg$ can be easily encoded as an acyclic WFSA $\fsa$ with $\abs{\fsa}=\abs{\cfg}$, by creating a weighted path of length $k$ and weight $w$\footnote{\label{fn:encodecfg}For example, the production $\edge{\nt{S}}{\nt{NP}\ \nt{VP}}$ would be encoded as a path of length 3 accepting the sequence $\nt{NP}\,\nt{VP}\,\hat{\nt{S}}$.  The production's weight may arbitrarily be placed on the first arc of the path, the other arcs having weight $\one$ (see \cref{app:semiring}).\looseness=-1} for each CFG production of size $k$ and weight $w$, terminating in a final state, and then merging the initial states of these paths into a single state that becomes the initial state of the resulting WFSA. The paths are otherwise disjoint. Importantly, this WFSA can then be determinized and minimized \cite{mohri-1997-finite} to potentially reduce the number of states and arcs (while preserving the total weight of each sequence) and thus speed up parsing \cite{klein2001time}. Among other things, this will merge common prefixes and common suffixes.

In general, however, the grammar can be specified by any WFSA $\fsa$---not necessarily deterministic.  This could be compiled from weighted regular expressions, or be an encoded Markov model trained on observed productions \citep{collins-1999}, or be obtained by merging states of another WFSA grammar \citep{stolcke-omohundro-1994-hmm} in order to smooth its weights and speed it up.

The WFSA has states $\dfaStates$ and weighted arcs (or edges) $\fsaEdges$, over an alphabet $\dfaAlpha$ consisting of $\terminals\cup\nonterminals$ together with hatted nonterminals like $\hat{A}$.  Its initial and final states are denoted by $\dfaStart\subseteq\dfaStates$ and $\dfaFinish\subseteq\dfaStates$, respectively.\footnote{Note that if the WFSA is obtained as described above, it will only have one initial state.}  We denote an arc of the WFSA by $(\dfaEdge{q}{a}{q'})\in\fsaEdges$ where $q,q'\in\dfaStates$ and $a\in\dfaAlpha\cup\{\emptystring\}$.  This corresponds to an axiom with the same weight as the edge. $q \in \dfaStart$ corresponds to an axiom whose weight is the initial-state weight of $q$.  
The item $q \in \dfaFinish$ is true not only if $q$ is a final state but more generally if $q$ has an $\emptystring$-path of length $\geq 0$ to a final state; the item's weight is the total weight of all such $\emptystring$-paths, where a path's weight includes its final-state weight.

For a state $q\in\dfaStates$ and symbol $A\in\nonterminals$, the precomputed side condition $\dfaEdge{q}{A}{\star}$ is true iff there exists a state $q'\in\dfaStates$ such that 
$\dfaEdge{q}{A}{q'}$ exists in $\fsaEdges$.
Additionally, the precomputed side condition $\dfaEdge{q}{*\hat{A}}{\star}$ is true if there exists a path starting from $q$ that eventually reads $\hat{A}$.  As these are only used as side conditions, they may be given any non-$\zero$ weight. \looseness=-1

The \earleyFSA deduction system is given in \cref{tab:fsa}.  It can be run in time $\bigo{\nN^3\abs{\fsa}}$.
It is similar to \earleyDeductionFast, where the dotted rules have been replaced by WFSA states.  However, unlike a dotted rule, a state does not specify a {\predict}ed left-hand-side nonterminal.  As a result, when any deduction rule ``advances the dot'' to a new state $q$, it builds a provisional item $\state{j,k,q?}$ that is annotated with a question mark.  This mark represents the fact that although $q$ is compatible with several left hand sides $A$ (those for which $\dfaEdge{q}{*\hat{A}}{\star}$ is true), the left context $\spanx{0:j}$ might not call for any of those nonterminals.  If it calls for at least one such nonterminal $A$, then the new \filter rule will remove the question mark, allowing further progress.

One important practical advantage of this scheme for natural language parsing is that it prevents a large-vocabulary slowdown.\footnote{\citet{earley70} used 1-word lookahead for this; see \cref{app:lookahead}.}
In \earleyDeduction, applying \predict to (say) $\state{3,4,\edge{\nt{NP}}{\nt{Det}\ \bigcdot\ \nt{N}}}$ results in thousands of items of the form $\state{4,4,\edge{\nt{N}}{\bigcdot\ a}}$ where $a$ ranges over all nouns in the vocabulary.  But \earleyFSA in the corresponding situation will predict only $\state{4,4,q}$ where $q$ is the initial state, without yet predicting the next word.  If the next input word is $\state{4,5,\term{happy}}$, then \earleyFSA follows just the $\term{happy}$ arcs from $q$, yielding items of the form $\state{4,5,q'?}$ (which will then be \filter{}ed away since $\term{happy}$ is not a noun).

Note that $\scan$, $\compone$ and $\comptwo$ are ternary, rather than binary as in $\earleyDeductionFast$. For further speed-ups we can apply the fold transform on these rules in a similar manner as before, resulting in binary deduction rules. We present this binarized version in \cref{app:fsa-binarized}.

As before, we must eliminate unary and nullary rules before parsing; \cref{app:fsa-null} explains how to do this with a WFSA grammar.
In addition, although \cref{tab:fsa} allows the WFSA to contain $\emptystring$-arcs, \cref{app:fsa-null} explains how to eliminate $\emptystring$-\emph{cycles} in the WFSA, which could prevent us from converging, for the usual reason that an item $\state{i,j,q}$ could participate in its own derivation.  Afterwards, there is again a nearly acyclic order in which the deduction engine can prove items (as in \cref{app:twopass} or \cref{app:pseudocode-pqueue}).

As noted above, we can speed up \earleyFSA by reducing the size of the WFSA.  Unfortunately, minimization of general FSAs is NP-hard.  However, we can at least seek the minimal \emph{deterministic} WFSA $\fsa'$ such that $\abs{\fsa'}\le\abs{\fsa}$, at least in most semirings \cite{mohri-2000-tcs,eisner-2003-simpler}.  The determinization \citep{aho1986compilers} and minimization \citep{aho1974design, revuz92} algorithms for the boolean semiring are particularly well-known.  Minimization merges states, which results in merging items, much as when \earleyDeductionFast merged items that had different pre-dot symbols \citep{leermakers92, Nederhof1997AVO, moore-2000-improved}.

Another advantage of the WFSA presentation of Earley's is that it makes it simple to express a tighter bound on the runtime.  Much of the grammar size $\abs{\cfg}$ or $\abs{\fsa}$ is due to terminal symbols that are not used at most positions of the input.  Suppose the input is an ordinary sentence (one word at each position, unlike the lattice case in \cref{fn:lattice}), and suppose $c$ is a constant such that no state $q$ has more than $c$ outgoing arcs labeled with the same terminal
$a \in \terminals$.  Then when $\scan$ tries to extend $\state{i,j,q}$, it considers at most $c$ arcs.  Thus, the $\bigo{\abs{\fsa}}$ factor in our runtime (where $\abs{\fsa}=\abs{\fsaEdges}$) can be replaced with $\bigo{\abs{\dfaStates}\cdot c+\abs{\fsaEdges_{\nonterminals}}}$,
where 
$\fsaEdges_{\nonterminals}\subseteq\fsaEdges$ is the set of edges that are \emph{not} labeled with terminals.

\section{Practical Runtime of Earley's}\label{sec:exp} 
We empirically measure the runtimes of
\earleyDeduction, \earleyDeductionFast, and \earleyFSA.
We use the tropical semiring to find the highest-weighted derivation trees.
We use two grammars that were extracted from the PTB: Markov-order-2 (M2) and Parent-annotated Markov-order-2 (PM2).\footnote{Available at \url{https://code.google.com/archive/p/bubs-parser/}. M2 contains $52,\!009$ preterminal rules and $13,\!893$ other rules.
PM2 contains $52,\!009$ preterminal rules and $25,\!919$ other rules.
The downloaded grammars did not have nullary rules or unary chains.}
For each grammar, we ran our parsers
\citep[using the tropical semiring;][]{pin-tropical-1998}\saveforcameraready{} on $100$ randomly selected sentences of $5$ to $40$ words from the PTB test-set (mean $21.4$, stdev $10.7$), although we omitted sentences of length $> 25$ from the \earleyDeduction graph as it was too slow ($> 3$ minutes per sentence).
The full results are displayed in \cref{app:experiment}.
The graph shows that \earleyDeductionFast is roughly $20\times$ faster at all sentence lengths.
We obtain a further speed-up of $2.5\times$ by switching to \earleyFSA.

\section{Conclusion}
In this reference work, we have shown how the runtime of Earley's algorithm is reduced to $\bigo{\nN^3\abs{\cfg}}$ from the naive $\bigo{\nN^3\abs{\cfg}\abs{\rules}}$.
We presented this dynamic programming algorithm as a deduction system, which splits prediction and completion into two steps each, in order to share work among related items.
To further share work, we generalized Earley's algorithm to work with a grammar specified by a weighted FSA. We demonstrated that these speed-ups are effective in practice. We also provided details for efficient implementation of our deduction system.
We showed how to generalize these methods to semiring-weighted grammars by correctly transforming the grammars to eliminate cyclic derivations. We further provided a method to compute the total weight of all sentences with a given prefix under a semiring-weighted CFG.

We intend this work to serve as a clean reference for those who wish to efficiently implement an Earley-style parser or develop related incremental parsing methods.  For example, our deduction systems could be used as the starting point for 
\begin{itemize}[nosep]
\item neural models of incremental processing, in which each derivation of an item contributes not only to its weight but also to its representation in a vector space \citep[cf.][]{drozdov-etal-2019-unsupervised-latent,mei-et-al-2020-icml};
\item biasing an autoregressive language model toward high-weighted grammatical prefixes via product-of-experts decoding \citep[cf.][]{semanticmachines-2021-emnlp, roy-2022-benchclamp, fang-2022-whole-truth};
\item extensions to incremental parsing of more or less powerful grammar formalisms.  
\end{itemize}

\section{Limitations}

Orthogonal to the speed-ups discussed in this work, \citet{earley70}
described an extension that we do not include here, which filters
deduction items using $k$ words of lookahead.  (However, we do treat 1-word
lookahead and left-corner parsing in \cref{app:lookahead}.)

While our deduction system runs in time proportional to the grammar
size $\abs{\cfg}$, this size is measured only after unary and
nullary productions have been eliminated from the grammar---which
can increase the grammar size as discussed in \cref{app:unary,app:null}.

We described how to compute prefix weights only for
\earleyDeductionFast, and we gave a prioritized execution scheme (\cref{app:pseudocode-pqueue}) only for \earleyDeductionFast. The versions for \earleyFSA should be similar.

Computing sentence weights \labelcref{eq:z-input} and prefix weights
\labelcref{eq:prefix-weight} involves a sum over infinitely many
trees.  In arbitrary semirings, there is no guarantee that such sums
can be computed.  Computing them requires summing geometric series
and---more generally---finding minimal solutions to systems of
polynomial equations.  See discussion in \cref{app:semiring} and
\cref{app:null}.  Non-commutative semirings also present special challenges; see \cref{app:noncomm}.

\section*{Acknowledgments}

We thank Mark-Jan Nederhof for useful references and criticisms,
and several anonymous reviewers for their feedback.  Any remaining
errors are our own.

Andreas Opedal is supported by the Max Planck ETH Center for Learning Systems.

\bibliography{acl2023}
\bibliographystyle{acl_natbib}

\appendix

\section{Semirings}\label{app:semiring}
As mentioned in \cref{sec:cfg}, the definition of weighted context-free grammars rests on the definition of semirings.
A semiring $\semiring$ is a 5-tuple $\semiringtuple$, where the set $\semiringtype$ is equipped with two operators: $\oplus$, which is associative and commutative, and $\otimes$, which is associative and distributes over $\oplus$.
The semiring contains values $\zero,\one\in\semiringtype$ such that $\zero$ is an identity element for $\oplus$ ($w \oplus \zero = \zero \oplus w = w, \forall w \in \semiringtype$) and annihilator for $\otimes$ ($w \otimes \zero = \zero \otimes w = \zero, \forall w \in \semiringtype$) and $\one$ is an identity for $\otimes$ ($w \otimes \one = \one \otimes w = w, \forall w \in \semiringtype$).

A semiring is commutative if additionally $\otimes$ is commutative. A closed semiring has an additional operator $*$ satisfying the axiom $(\forall w \in \semiringtype)\; w^* = \one \oplus w \otimes w^* = \one \oplus w^* \otimes w$.  The interpretation is that $w^*$ returns the infinite sum $\one \oplus w \oplus (w \otimes w) \oplus (w \otimes w \otimes w) \oplus \cdots$.

As an example that may be of particular interest, \citet{goodman-1999-semiring} shows how to construct a (non-commutative) derivation semring, so that $\Z_\vx$ in \cref{eq:z-input} gives the best derivation (parse tree) along with its weight, or alternatively a representation of the forest of all weighted derivations.  This is how a weighted recognizer can be converted to a parser.

\section{Earley's Original Algorithm as a Deduction System}\label{app:earley-original} 

\Cref{sec:earley} introduced the deduction system  that corresponds to Earley's original algorithm.  We explain and analyze it here.
Overall, the three rules of this system, \earleyDeduction (\cref{tab:systems}), correspond to possible steps in a top-down recursive descent parser \citep{aho1986compilers}:

\begin{itemize}[noitemsep]
\item \scan consumes the next single input symbol (the base case of recursive descent);
\item \predict calls a subroutine to consume an entire constituent of a given nonterminal type by recursively consuming its subconstituents;
\item \complete returns from that subroutine.  
\end{itemize}
How then does it differ from recursive descent?  Rather like depth-first search, Earley's algorithm uses memoization to avoid redoing work, which avoids exponential-time backtracking and infinite recursion.  But like breadth-first search, it pursues possibilities in parallel rather than by backtracking.  The steps are invoked not by a backtracking call stack but by a deduction engine, which can deduce new items in any convenient order.
The effect on the recursive descent parser is essentially to allow co-routining \citep{knuth97}: execution of a recursive descent subroutine can suspend until further input becomes available or until an ancestor routine has returned and memoized a result thanks to some other nondeterministic execution path.

\subsection{Predict}\label{sec:predict}

To look for constituents of type $B$ starting at position $j$, using the rule $\edge{B}{\rho}$, we need to prove $\prededstate$.
Earley's algorithm imposes $\predstate$ as a side condition, so that we only start looking if such a constituent $B$ could be combined with some item to its left.\footnote{\citet{minnen-1996} and \citet{eisner-blatz-2007} explain that this side condition is an instance of the ``magic sets'' technique that filters some unnecessary work from a bottom-up algorithm \citep{ramakrishnan-1991}.} 

\begin{prooftree}
\AxiomC{$\edge{B}{\rho}$}
\LeftLabel{\smaller \pred:}
\RightLabel{$\predstate$}
\UnaryInfC{$\prededstate$}
\end{prooftree}

\paragraph{Runtime analysis.}
How many ways are there to jointly instantiate the two antecedents of \pred with actual items?  The pair of items is determined by making four choices:\footnote{Treating these choices as free and independent is enough to give us an upper bound.  In actuality, the choices are not quite independent---for example, any provable item has $i \leq j$---but there are no interdependencies that could be exploited to tighten our asymptotic bound.} indices $i$ and $j$ with a domain size of $\nN+1$, dotted production $\edge{A}{\mu \bigcdot B\, \nu}$ with domain size $\abs{\cfg}$, and production $\edge{B}{\rho}$ with a domain size of $\abs{\rules}$.
Therefore, the number of instantiations of \pred is $\bigo{\nN^2\abs{\cfg}\abs{\rules}}$.  That is then \pred's contribution to the runtime of a suitable implementation of forward chaining deduction, using Theorem 1 of \citet{mcAllester02}.\footnote{Technically, that theorem also directs us to count the instantiations of just the first antecedent, namely $\bigo{\abs{\rules}}$.  But this term can be ignored, as it is dominated in the asymptotic analysis by the number of complete instantiations $\bigo{\nN^2\abs{\cfg}\abs{\rules}}$.  In general, we can stick to upper-bounding the number of complete instantiations whenever this upper bound treats the choices as independent, since then it always equals or exceeds the number of partial instantiations.}

\subsection{Scan}\label{sec:scan}%
If we have proved an incomplete item $\scanstate$, we can advance the dot if the next terminal symbol is $a$:
\begin{prooftree}
\def\defaultHypSeparation{\hspace{0pt}}
\AxiomC{$\scanstate$}
\AxiomC{$\state{j, k, a}$}
\LeftLabel{\smaller \scan:}
\BinaryInfC{$\scannedstate$}
\end{prooftree}
This makes progress toward completing the $A$. Note that \scan pushes the antecedent to a subsequent item set $\stateset{k}$. Since terminal symbols have a span width of $1$, it follows that $j=k-1$.

\paragraph{Runtime analysis.}
\scan has three free choices: indices $i$ and $j$ with a domain size of $\nN+1$, and dotted production $\edge{A}{\mu \bigcdot B\, \nu}$ with domain size $\abs{\cfg}$.
Therefore, \scan contributes  $\bigo{\nN^2\abs{\cfg}}$ to the overall runtime.

\subsection{Complete}\label{sec:comp}

Recall that having $\state{i,j,\edge{A}{\mu \bigcdot B\, \nu}}$ allowed us to start looking for a $B$ at position $j$ (\pred).  Once we have found a complete $B$ by deriving $\compstate$, we can advance the dot in the former rule:
\begin{prooftree}
\def\defaultHypSeparation{\hspace{0pt}}
\AxiomC{$\predstate$}
\AxiomC{$\compstate$}
\LeftLabel{\smaller \comp:}
\BinaryInfC{$\compedstate$}
\end{prooftree}

\paragraph{Runtime analysis.}
\comp has five free choices: indices $i$, $j$, and $k$ with a domain size of $\nN+1$, dotted production $\edge{A}{\mu \bigcdot B\, \nu}$ with domain size $\abs{\cfg}$, and the complete production $\edge{B}{\rho}$ with a domain size of $\abs{\rules}$.
Therefore, \comp contributes $\bigo{\nN^3\abs{\cfg}\abs{\rules}}$ to the runtime.

\subsection{Total Space and Runtime}

By a similar analysis of free choices, the number of items that the \earleyDeduction deduction system will be able to prove is $\bigo{\nN^2\abs{\cfg}}$.  This is a bound on the space needed by the forward chaining implementation to store the items that have been proved so far and index them for fast lookup \cite{mcAllester02,eisner-etal-2005-compiling,eisner-2023-tacl}.  

Following Theorem 1 of \citet{mcAllester02}, adding this count to the total number of rule instantiations from the above sections yields a bound on the total runtime of the \earleyDeduction algorithm, namely $\bigo{\nN^3\abs{\cfg}\abs{\rules}}$ as claimed.  

\section{Previous Speed-ups}\label{sec:past}

We briefly discuss past approaches used to improve the asymptotic efficiency of \earleyDeduction.

\citet{leermakers92} noted that in an item of the form $\stdstate$, the sequence $\mu$ is irrelevant to subsequent deductions.  Therefore, he suggested (in effect) replacing $\mu$ with a generic placeholder $\star$.  This merges items that had only differed in their $\mu$ values, so the algorithm processes fewer items.
This technique can also be seen in \citet{moore-2000-improved} and \citet{klein2001, klein2001time}.
Importantly, this means that each nonterminal only has one complete item, $\closestate$, for each span.
This effect alone is enough to improve the runtime of Earley's to $\bigo{\nN^3\abs{\cfg}+\nN^2\abs{\cfg}\abs{\rules}}$.  
Our \cref{sec:new-comp} will give a version of the trick that only gets this effect, by folding the \complete rule.  The full version of \citet{leermakers92}'s trick is subsumed by our generalized approach in \cref{sec:fsa}.\looseness=-1

While the GHR algorithm---a modified version of Earley's algorithm---is commonly known to be $\bigo{\nN^3\abs{\cfg}\abs{\rules}}$, \citet[\S3]{grahamHR80} provide a detailed exploration of the low-level implementation of their algorithm that enables it to be run in $\bigo{\nN^3\abs{\cfg}}$ time.
This explanation spans 20 pages and includes techniques similar to those mentioned in \cref{sec:new}, as well as discussion of data structures.  To the best of our knowledge, these details have not been carried forward in subsequent presentations of GHR \citep{stolcke95, goodman-1999-semiring}.
In the deduction system view, we are able to achieve the same runtime quite easily and transparently by folding 
both \complete (\cref{sec:new-comp}) and \predict (\cref{sec:new-pred}).%
In both cases, this eliminates the pairwise interactions between all $\abs{\cfg}$ dotted productions and all $\abs{\rules}$ complete productions, thereby reducing $\abs{\cfg}\abs{\rules}$ to $\abs{\cfg}$.

\section{Correspondence Between \earleyDeduction and \earleyDeductionFast}\label{app:one-to-one}

The proofs of \earleyDeductionFast are in one-to-one correspondence with the proofs of \earleyDeduction.

We show the key steps in transforming between the two styles of proof.  
\Cref{tab:pred-proof} shows the correspondence between an application of \pred and an application of \predone and \predtwo, while \cref{tab:comp-proof} shows the correspondence between an application of \comp and an application of \compone and \comptwo.

\renewcommand{\arraystretch}{1}
\setlength{\tabcolsep}{6pt}

\begin{table*}[t]
\centering
\small
\begin{tabularx}{0.9\linewidth}{C C}
    \earleyDeduction & \earleyDeductionFast \\ \toprule 
    \hspace{10pt}
        \def\defaultHypSeparation{\hspace{0pt}}
        \AxiomC{$\edge{B}{\nu}$}
        \RightLabel{$\predstate$}
        \LeftLabel{\smaller \pred:}
        \UnaryInfC{$\prededstate$}
        \DisplayProof 
    &
        \def\defaultHypSeparation{\hspace{0pt}}

        \AxiomC{}
        \RightLabel{$\predstate$}
        \LeftLabel{\smaller \predone:}
        \UnaryInfC{$\fillstate$}
        \DisplayProof

        \AxiomC{$\edge{B}{\nu}$}
        \LeftLabel{\smaller \predtwo:}
        \RightLabel{$\fillstate$}
        \UnaryInfC{$\prededstate$}
        \DisplayProof
\end{tabularx}
\caption{Any application of \pred in \earleyDeduction has a one-to-one correspondence with an application of \predone and \predtwo in \earleyDeductionFast.
Note that it is not possible for a derivation in \earleyDeductionFast to have 
$\prededstate$
without state 
$\fillstate$ and an application of \predone.
}
\label{tab:pred-proof}
\end{table*}

\renewcommand{\arraystretch}{1}
\setlength{\tabcolsep}{6pt}

\begin{table*}[t]
\centering
\small
\begin{tabularx}{0.97\linewidth}{C C}
 \earleyDeduction & \earleyDeductionFast \\ \toprule
    \hspace{0pt}
        \def\defaultHypSeparation{\hspace{0pt}}
        \AxiomC{$\predstate$}
        \AxiomC{$\compstate$}
        \LeftLabel{\smaller \comp:}
        \BinaryInfC{$\compedstate$}
        \DisplayProof
    &
        \def\defaultHypSeparation{\hspace{0pt}}
        \AxiomC{$\predstate$}
        \AxiomC{$\compstate$}
        \LeftLabel{\smaller \compone:}
        \UnaryInfC{$\closestate$}
        \LeftLabel{\smaller \comptwo:}
        \BinaryInfC{$\compedstate$}
        \DisplayProof 
\end{tabularx}
\caption{Any application of \comp in \earleyDeduction has a one-to-one correspondence with an application of \compone and \comptwo in \earleyDeductionFast.
Note that it is not possible for a derivation in \earleyDeductionFast to have $\closestate$ without state $\compstate$ and an application of \compone.
}
\label{tab:comp-proof}
\end{table*}

\section{Eliminating Unary Cycles}\label{app:unary}

As mentioned in \cref{sec:weighted}, our weighted deduction system requires that we eliminate unary cycles from the grammar. \citet[\S4.5]{stolcke95} addresses the problem of unary production cycles by modifying the deduction rules.\footnote{\citet{johnsonSoftware} provides an implementation of CKY (and the inside-outside algorithm) that allows unary productions and handles unary cycles in a similar way.} He assumes use of the probability semiring, where $\semiringtype=[0,1]$, $\oplus=+$, and $\otimes=\times$. In that case, inverting a single $\abs{\nonterminals} \times \abs{\nonterminals}$ matrix suffices to compute the total weight of all rewrite sequences $\rewrites{A}{B}$, known as \defn{unary chains}, for each ordered pair $A,B \in \nonterminals^2$.\footnote{In a PCFG in which all rule weights are $> 0$, this total weight is guaranteed finite provided that all nonterminals are generating (\cref{fn:coacc}).}  His modified rules then ignore the original unary productions and refer to these weights instead.

We take a very similar approach, but instead describe it as a
transformation of the weighted grammar, leaving the deduction system
unchanged.  We generalize from the probability semiring to any closed
semiring---that is, any semiring that provides an operator $*$
to compute geometric series sums in closed form (see \cref{app:semiring}).
In addition, we improve the construction: we do not collapse all unary chains as \citet{stolcke95} does, but only those subchains that can appear on cycles.  This prevents the grammar size from blowing up more than necessary (recall that the parser's runtime is proportional to grammar size).  For example, if the unary productions are $\edge{A_i}{A_{i+1}}$ for all $1 \leq i < \nK$, then there is no cycle and our transformation leaves these $\nK-1$ productions unchanged, rather than replacing them with $\nK(\nK-1)/2$ new unary productions that correspond to the possible chains $\rewrites{A_i}{A_j}$ for $1 \leq i \leq j \leq \nK$.\saveforcameraready{}

Given a weighted CFG $\cfg = \tuple{\nonterminals, \terminals, \rules, \start, w}$, consider the weighted graph whose vertices are $\nonterminals$ and whose weighted edges $\edge{A}{B}$ are given by the unary productions $\edge{A}{B}$. (This graph may include self-loops such as $\edge{A}{A}$.)  Its strongly connected components (SCCs) will represent unary production cycles and can be found in linear time (and thus in $\bigo{\abs{\cfg}}$ time).  For any $A$ and $B$ in the same SCC, $w(\rewrites{A}{B}) \in \semiringtype$ denotes the total weight of all rewrite sequences of the form $\rewrites{A}{B}$ (including the 0-length sequence with weight $\one$, if $A=B$).  For an SCC of size $\nK$, there are $\nK^2$ such weights and they can be found in total time $\bigo{\nK^3}$ by the Kleene--Floyd--Warshall algorithm \cite{lehmann77, tarjan81a, tarjan81b}.  In the real semiring, this algorithm corresponds to using Gauss-Jordan elimination to invert $I-E$, where $E$ is the weighted adjacency matrix of the SCC (rather than of the whole graph as in \citet{stolcke95}).  In the general case, it computes the infinite matrix sum $I \oplus E \oplus (E \otimes E) \oplus \cdots$ in closed form, with the help of the $^*$ operator of the closed semiring.

We now construct a new grammar $\cfg'=\tuple{\nonterminals',\terminals,\rules',\bar{\start}, w'}$ that has no unary cycles, as follows.
For each $A\in\nonterminals$, our $\nonterminals'$ contains two nonterminals, $\bar{A}$ and $\underline{A}$.  For each ordered pair of nonterminals $A, B \in \nonterminals^2$ that fall in the same SCC, $\rules'$ contains a production $\edge{\bar{A}}{\underline{B}}$ with 
$\weightprime{\edge{\bar{A}}{\underline{B}}} = \weight{\rewrites{A}{B}}$.  For every rule $\edge{A}{\rho}$ in $\rules$ that is \emph{not} of the form $\edge{A}{B}$ where $A$ and $B$ fall in the same SCC, $\rules'$ also contains a production $\edge{\underline{A}}{\bar{\rho}}$ with 
$\weightprime{\edge{\underline{A}}{\bar{\rho}}}=\weight{\edge{A}{\rho}}$,
where $\bar{\rho}$ is a version of $\rho$ in which each nonterminal $B$ has been replaced by $\bar{B}$.   Finally, as a constant-factor optimization, $\bar{A}$ and $\underline{A}$ may be merged back together if $A$ formed a trivial SCC with no self-loop: that is, remove the weight-$\one$ production $\edge{\bar{A}}{\underline{A}}$ from $\rules'$ and replace all copies of $\bar{A}$ and $\underline{A}$ with $A$ throughout $\cfg'$.

\saveforcameraready{}

Of course, as \citet{aycock-2002-practical} noted, this grammar transformation does change the derivations (parse trees) of a sentence, which is also true for the grammar transformation in \cref{app:null} below.  A derivation under the new grammar (with weight $w$) may represent infinitely many derivations under the old grammar (with total weight $w$).   In principle, if the old weights were in the derivation semiring (see \cref{app:semiring}), then $w$ will be a representation of this infinite set.  This implies that the $*$ operator in this section, and the polynomial system solver in \cref{app:null} below, must be able to return weights in the derivation semiring that represent infinite context-free languages.

\section{Eliminating Nullary Productions}\label{app:null}

In addition to unary cycles (\cref{app:unary}) we must eliminate nullary productions in order to avoid cyclic proofs, as mentioned in \cref{sec:weighted}. 
This must be done \emph{before} eliminating unary cycles, since eliminating nullary productions can create new unary productions.  \citet[\S7.1.3]{hopcroft} explain how to do this in the unweighted case.
\Citet[\S4.7.4]{stolcke95} sketches a generalization to the probability semiring, but it also uses the non-semiring operations of division and subtraction (and is not clearly correct). We therefore give an explicit general construction.

While we provide a method that handles nullary productions by modifying the grammar, it is also possible to instead modify the algorithm to allow advancing the dot over \defn{nullable} nonterminals, i.e., nonterminals $A$ such that the grammar allows $\rewrites{A}{\emptystring}$ \citep{aycock-2002-practical}.

Our first step, like \citeauthor{stolcke95}'s, is to compute the ``null weight'' 
\begin{align}\label{eq:nullweight}
\e{A} &\defeq \weight{\rewrites{A}{\emptystring}} \defeq \smashoperator{\bigoplus_{\substack{\tree\in\subtrees{A}: \\ \mathrm{yield}(T)=\emptystring}}} \weight{\tree}
\end{align}
for each $A \in \nonterminals$. 
Although a closed semiring does not provide an operator for this summation, these values are a solution to the system of $\abs{\nonterminals}$ polynomial equations\footnote{If $(\edge{A}{\emptystring})\in\rules$, it will be covered by the case $n=0$.}
\begin{align}
    \e{A} &= 
    \smashoperator{\bigoplus_{(\edge{A}{B_1 \cdots B_n})\in\rules}} \weight{\edge{A}{B_1 \cdots B_n}} \otimes \bigotimes_{i=1}^{n} \e{B_i} \label{eq:nullsystem}
\end{align}
In the same way, the free weights from \cref{eq:free-weight} in
\cref{sec:prefix} are a solution to the system
\begin{align}
    \freeweight{A} &= 
    \smashoperator{\bigoplus_{(\edge{A}{\rho})\in\rules}}
        \weight{\edge{A}{\rho}} \otimes \bigotimes_{i:
        \rho_i \in \nonterminals} \freeweight{\rho_i} \label{eq:freesystem}
\end{align}
which differs only in that $\rho$ is allowed to contain terminal
symbols.  In both cases, the distributive property of semirings is
being used to recursively characterize a sum over what may be
infinitely many trees.  A solution to system~\labelcref{eq:nullsystem}
must exist for the sums in \cref{eq:z-input} to be well-defined in the
first place.  (Similarly, a solution to
system~\labelcref{eq:freesystem} must exist for the sums in
\cref{eq:prefix-weight,eq:free-weight} to be well-defined.)  If there
are multiple solutions, the desired sum is given by the ``minimal''
solution, in which as many variables as possible take on value
$\zero$.  Often in practice the minimal solution can be found using
fixed-point iteration, which initializes all free weights to $\zero$
and then iteratively recomputes them via
system~\labelcref{eq:nullsystem} (respectively
system~\labelcref{eq:freesystem}) until they no longer change (e.g.,
at numerical convergence).  For example, this is guaranteed to work in
the tropical semiring
$(\semiringtype,\oplus,\otimes,\zero,\one) = (\mathbb{R}_{\geq
  0},\min,+,\infty,0)$
and more generally in $\omega$-continuous semirings under conditions given by \citet{kuich97semirings}.  \Citet{esparza-2007-extension-of-newton} and \citet{etessami-yannakakis-2009-monotone-systems} examine a faster approach based on Newton's method.  \Citet{nederhof-satta-2008} review methods for the case of the real weight semiring
$(\semiringtype,\oplus,\otimes,\zero,\one) = (\mathbb{R}_{\geq
  0},+,\times,0,1)$.

\newcommand{\noneps}[1]{#1_{\neq\emptystring}}
Given the null weights $\e{A} \in \semiringtype$, we now modify the grammar as follows.  We adopt the convention that for a production $\edge{A}{\rho}$ that is not yet in $\rules$, we consider its weight to be $w(\edge{A}{\rho})=\zero$, and increasing this weight by any non-$\zero$ amount adds it to $\rules$.  For each nonterminal $B$ such that $\e{B}\neq\zero$, let us assume the existence of an auxiliary nonterminal $\noneps{B}\notin\nonterminals$ such that
$\notrewrites{\noneps{B}}{\emptystring}$ but
$\forall\vx\neq\emptystring$, $\weight{\rewrites{\noneps{B}}{\vx}}=\weight{\rewrites{B}{\vx}}$.   We iterate this step: as long as we can find a production $\edge{A}{\mu\, B\, \nu}$ in $\rules$ such that $\e{B} \neq \zero$, we modify it to the more restricted version $\edge{A}{\mu\, \noneps{B}\, \nu}$ (keeping its weight), but to preserve the possibility that $\rewrites{B}{\emptystring}$, we also increase the weight of the shortened production $\edge{A}{\mu\, \nu}$ by $\e{B}\otimes\weight{\edge{A}{\mu\, B\, \nu}}$.  

A production $\edge{A}{\rho}$ where $\rho$ includes $k$ nonterminals $B$ with $\e{B} \neq \zero$ will be gradually split up by the above procedure into $2^k$ productions, in which each $B$ has been either specialized to $\noneps{B}$ or removed.  The shortest of these productions is $\edge{A}{\emptystring}$, whose weight is $\weight{\edge{A}{\emptystring}}=\e{A}$ by \cref{eq:nullsystem}.  
So far we have preserved all weights $\weight{\rewrites{A}{\vx}}$, provided that the auxiliary nonterminals behave as assumed. For each $A$ we must now remove $\edge{A}{\emptystring}$ from $\rules$, and since $A$ can no longer rewrite as $\emptystring$, we rename all other rules $\edge{A}{\rho}$ to $\edge{\noneps{A}}{\rho}$. This closes the loop by defining the auxiliary nonterminals as desired.  

Finally, since $S$ is the start symbol, we add back $\edge{S}{\emptystring}$ (with weight $\e{S}$) as well as adding the new rule $\edge{S}{\noneps{S}}$ (with weight $\one$).
Thus (as in Chomsky Normal Form), the only nullary rule is now $\edge{S}{\emptystring}$, which may be needed to generate the 0-length sentence.  We now have a new grammar with nonterminals $\nonterminals' = \{S\} \cup \{\noneps{B}: B\in\nonterminals\}$.  To simplify the names, we can rename the start symbol $S$ to $\specialS$ and then drop the $\noneps{}$ subscripts.  Also, any nonterminals that \emph{only} rewrote as $\emptystring$ in the original grammar are no longer generating and can be safely removed (see \cref{fn:coacc}).

\renewcommand{\arraystretch}{2.6}

\begin{table*}[t]
\centering
\small
\begin{tabularx}{0.98\linewidth}{l l r l}
    & & & \\ \toprule 

    & \startrule: & $\prefixoutside(\state{0,0,\edge{B}{\,\bigcdot\, \star}}) $ & $\mathrel{\oplus}=\weight{\leftcorner{\start}{B}}$ \\
    & \predonemod: & $\prefixoutside(\fillstate) $ & $ \mathrel{\oplus}=\prefixoutside(\predstateC) \otimes \weight{\leftcorner{C}{B}} \otimes \incinside(\predstateC) $ \\
    & \predtwo: & $\prefixoutside(\prededstate) $ & $ \mathrel{\oplus}= \prefixoutside{\fillstate}$ \\
    & \scan: & $\prefixoutside(\scannedstate) $ & $ \mathrel{\oplus}= \prefixoutside{\scanstate}$ \\
    & \compone: & $\prefixoutside(\closestate) $ & $ \mathrel{\oplus}=\prefixoutside{\compstate}$ \\
    & \comptwo: & $\prefixoutside(\compedstate) $ & $ \mathrel{\oplus}= \prefixoutside{\prededstate}$ \\ 
    & \pos: & $\prefixoutside(\prefixstate) $ & $ \mathrel{\oplus}= \prefixoutside(\scannedstatep)\otimes \incinside(\scannedstatep) \otimes Z_\nu$ \\\bottomrule
\end{tabularx}
\caption{Explicit formulas for incrementing the prefix outside weights during one-step proofs for \earleyDeductionFast for the general case in which the grammar may be left-recursive, as explained in \cref{app:left-recursion}. Note that the prefix outside weights for \compone go unused for subsequent proof steps, and thus do not contribute to the prefix weights associated with the input string $\vx$. The prefix outside weight for $\prefixoutside(\prefixstate)$ is the desired prefix weight $\weight{\leftcorner{S}{\spanx{0:j}}}$.
}
\label{tab:prefix}
\vspace{-1pt}
\end{table*}

\section{Working With Left Corners}

\subsection{Recursive Chains in Prefix Outside Weights}{\label{app:left-recursion}}

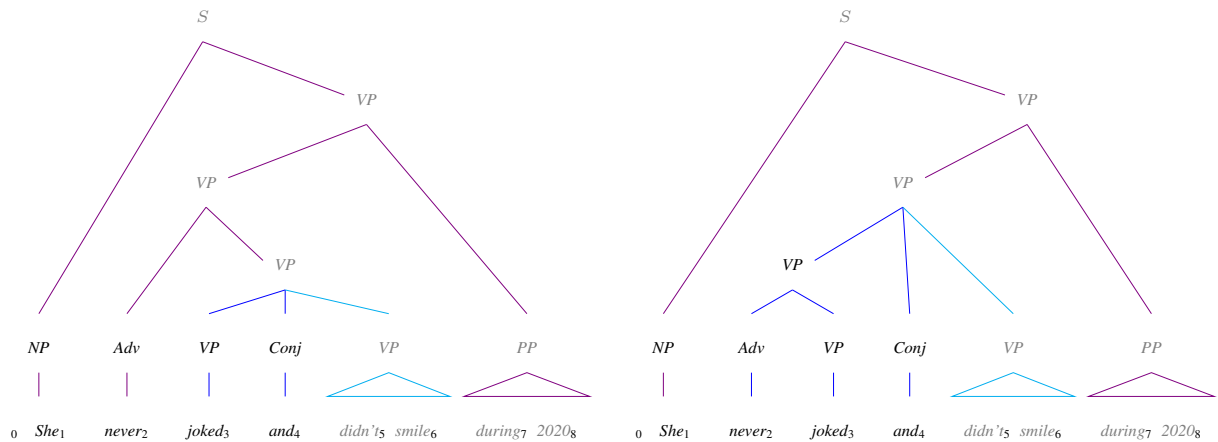
\begin{figure*}[t]
\centering
\small
\resizebox{\textwidth}{!}{%
\begin{forest}
[,phantom,s sep=.5cm, align=center
[
$\start$, color=\unseencolor
[$\nt{NP}$, tier=preterm, edge=\outsidecolor [\textsubscript{0}\quad $\term{She}$\textsubscript{1}, tier=word, edge=\outsidecolor] ]
[$\nt{VP}$, child anchor=west, color=\unseencolor, edge=\outsidecolor
[$\nt{VP}$, child anchor=east, edge=\outsidecolor, color=\unseencolor
[$\nt{Adv}$, tier=preterm, edge=\outsidecolor [$\term{never}$\textsubscript{2}, tier=word, edge=\outsidecolor]]
[$\nt{VP}$, calign=child, calign child=2, child anchor=west, edge=\outsidecolor, color=\unseencolor [$\nt{VP}$, tier=preterm, edge=\incinsidecolor [$\term{joked}$\textsubscript{3}, tier=word, edge=\incinsidecolor]] [$\nt{Conj}$, tier=preterm, edge=\incinsidecolor [$\term{and}$\textsubscript{4}, tier=word, edge=\incinsidecolor] ] [$\nt{VP}$, tier=preterm, color=\unseencolor, edge=\insidecolor [\textcolor{\unseencolor}{$\term{didn't}$}\textsubscript{5}\; \textcolor{\unseencolor}{$\term{smile}$}\textsubscript{6} ,roof, tier=word, edge=\insidecolor] ] ]
]
[$\nt{PP}$, tier=preterm, color=\unseencolor, edge=\outsidecolor [\textcolor{\unseencolor}{$\term{during}$}\textsubscript{7}\; \textcolor{\unseencolor}{$\term{2020}$}\textsubscript{8} ,roof, tier=word, edge=\outsidecolor ]]
]
]
[
$\start$, color=\unseencolor
[$\nt{NP}$, tier=preterm, edge=\outsidecolor [\textsubscript{0}\quad $\term{She}$\textsubscript{1}, tier=word, edge=\outsidecolor] ]
[$\nt{VP}$, child anchor=west, color=\unseencolor, edge=\outsidecolor
[$\nt{VP}$, child anchor=east, edge=\outsidecolor, color=\unseencolor
[$\nt{VP}$, child anchor=east, edge=\incinsidecolor [$\nt{Adv}$, edge=\incinsidecolor [$\term{never}$\textsubscript{2}, tier=word, edge=\incinsidecolor]] [$\nt{VP}$, tier=preterm, edge=\incinsidecolor [$\term{joked}$\textsubscript{3}, tier=word, edge=\incinsidecolor]] ]
[$\nt{Conj}$, edge=\incinsidecolor, tier=preterm [$\term{and}$\textsubscript{4}, tier=word, edge=\incinsidecolor] ]
[$\nt{VP}$, tier=preterm, color=\unseencolor, edge=\insidecolor [\textcolor{\unseencolor}{$\term{didn't}$}\textsubscript{5}\; \textcolor{\unseencolor}{$\term{smile}$}\textsubscript{6} ,roof, tier=word, edge=\insidecolor] ]
]
[$\nt{PP}$, tier=preterm, color=\unseencolor, edge=\outsidecolor [\textcolor{\unseencolor}{$\term{during}$}\textsubscript{7}\; \textcolor{\unseencolor}{$\term{2020}$}\textsubscript{8} ,roof, tier=word, edge=\outsidecolor ]]
]
]
]
\end{forest}
}
\caption{To find the prefix weight of the prefix ``She never joked and,'' the algorithm must consider \emph{all} derivations of \emph{all} complete sentences with that prefix (\cref{sec:prefix}).  This figure shows two derivations of one such completion---the ambiguous sentence ``She never joked and didn’t smile during 2020.'' The left derivation corresponds to a proof tree for which the unique incomplete item ending at position 4 is $\state{2, 4, \edge{\nt{VP}}{\nt{VP}\ \nt{Conj} \bigcdot \nt{VP}}}$ and the right derivation corresponds to a proof tree for which the unique incomplete item ending at position 4 is $\state{1, 4, \edge{\nt{VP}}{\nt{VP}\ \nt{Conj} \bigcdot \nt{VP}}}$.
We colorize the productions associated with these items' \textcolor{\incinsidecolor}{incomplete inside weight}, \textcolor{\insidecolor}{future inside weight} and \textcolor{\outsidecolor}{outside weight}. The prefix outside weight of $\prefixoutside(\state{4,4, \edge{\nt{VP}}{\bigcdot\ \star}})$ sums the product of these three weights over all derivations at \predone (\cref{eq:narrow-outside}).  It makes use of use of free weights to sum over all expansions of any nonterminals that were predicted to follow position 4: in these examples, $Z_{\nt{VP}}$  and $Z_{\nt{PP}}$ are included in, respectively, the future inside weight and the prefix outside weight of the incomplete antecedent item ($\state{2, 4, \edge{\nt{VP}}{\nt{VP}\ \nt{Conj} \bigcdot \nt{VP}}}$ or $\state{1, 4, \edge{\nt{VP}}{\nt{VP}\ \nt{Conj} \bigcdot \nt{VP}}}$).  
The example derivations shown both contribute one copy of $\edge{\nt{VP}}{\nt{VP}\ \nt{PP}}$ to the prefix outside weight. However, since that production is left-recursive, the prefix is also consistent with other completions that use it $k$ times to produce $k$ arbitrary future prepositional phrases, for any $k=0,1,2,\ldots$.  To sum over all such possibilities we provide a modified \predone in \cref{app:left-recursion}.  This summation over copies of $\edge{\nt{VP}}{\nt{VP}\ \nt{PP}}$ (and over the possible expansions of its future $\nt{PP}$) is needed both when $\nt{VP}$ is predicted at position 1 and again when $\nt{VP}$ is predicted at position 2.
}
\label{fig:prefix-tree}
\end{figure*}

As mentioned in \cref{sec:prefix}, there is a subtle issue that arises if the grammar has left-recursive productions. 
Consider the left-recursive rule $\lrrule$. Using \cref{eq:narrow-outside}, the prefix outside weight of the predicted item $\lrstate$ will only include the weight corresponding to one rule application of $\lrrule$, but correctness demands that we account for the possibility of recursively applying $\lrrule$ as well. A well-known technique to remove left-recursion is the left-corner transform \citep{rosenkrantz1970deterministic, johnson-roark-2000-compact}. As that may lead to drastic increases in grammar size, however, we instead provide a modification of \predone that deals with this technical complication (which adapts \citet[\S4.5.1]{stolcke95} to our improved deduction system and generalizes it to closed semirings). \cref{fig:prefix-tree} provides some further intuition on the left-recursion issue.

We require some additional definitions: $B$ is a left child of $A$ iff there exists a rule $\edge{A}{B \, \rho}$. The \emph{reflexive} and transitive closure of the left-child relation is $\leftcorner{}{}$, which was already defined in \cref{sec:cfg}. A nonterminal $A$ is said to be left-recursive if $A$ is a nontrivial left corner of itself, i.e., if $\transitiveleftcorner{A}{A}$ (meaning that $\edge{A}{B \, \rho}$ and $\leftcorner{B}{A}$ for some $B$).  A grammar is left-recursive if at least one of its nonterminals is left-recursive. 

To deal with left-recursive grammars, we collapse the weights of left-recursive paths similarly as we did with unary cycles (see \cref{app:unary}), and $\otimes$-multiply in at the \predone step.

We consider the left-corner multigraph: given a weighted CFG $\cfg = \tuple{\nonterminals, \terminals, \rules, \start, w}$, its vertices are $\nonterminals$ and its edges are given by the left-child relations, with one edge for every production. Each edge is associated with a weight equal to the weight of the corresponding production $\otimes$-times the free weights of the nonterminals on the right hand side of the production that are not the left-child. For instance, for a production $\edge{A}{B\ C\ D}$, the weight of the corresponding edge in the graph will be $\weight{\edge{A}{B\ C\ D}} \otimes \freeweight{C} \otimes \freeweight{D}$.  
This graph's SCCs represent the left-corner relations.
For any $A$ and $B$ in the same SCC $\weight{\leftcorner{A}{B}} \in \semiringtype$ denotes the total weight of all left-corner rewrite sequences of the form $\leftcorner{A}{B}$, including the free weights needed to compute the prefix outside weights. 
These can, again, be found in $\bigo{K^3}$ time with the Kleene--Floyd--Warshall algorithm \cite{lehmann77, tarjan81a, tarjan81b}, where $K$ is the size of the SCC. 
These weights can be precomputed and have no effect on the runtime of the parsing algorithm. We replace \predone with the following:

\begin{prooftree}
\AxiomC{}
\RightLabel{\makecell[c]{$\predstateC$ \\ $\mu\neq\emptystring$ \quad $\leftcorner{C}{B}$}}
\LeftLabel{\smaller \predonemod:}
\UnaryInfC{$\fillstate$}
\end{prooftree}

\noindent A one-step proof of \predonemod contributes
\begin{align}
  & \prefixoutside(\predstateC) \otimes \weight{\leftcorner{C}{B}} \nonumber \\
  & \mbox{} \otimes \incinside(\predstateC) \label{eq:pred1lr-outside}
\end{align}
to the prefix outside weight $\prefixoutside(\prededstate)$.

Note that the case $B=C$ recovers the standard \predone, and such rules will always be instantiated since $\leftcorner{}{}$ is reflexive.
The \predonemod rule has three side conditions (whose visual layout here is not significant). Its consequent will feed into \predtwo; the condition $\mu\neq\emptystring$ ensures that the output of \predtwo cannot serve again as a side condition to \predone, since the recursion from $C$ was already fully computed by the $\leftcorner{C}{B}$ item.  However, since this condition prevents \predonemod from predicting anything at the start of the sentence, we must also replace the start axiom $\state{0,0,\edge{S}{\,\bigcdot\, \star}}$ with a rule that resembles \predone and derives the start axiom together with all its left corners: 

\begin{prooftree}
\AxiomC{}
\RightLabel{$\leftcorner{\start}{B}$}
\LeftLabel{\smaller \startrule:}
\UnaryInfC{$\state{0,0,\edge{B}{\,\bigcdot\, \star}}$}
\end{prooftree}

The final formulas for aggregating the prefix outside weights are spelled out explicitly in \cref{tab:prefix}.  Note that we did not spell out a corresponding prefix weight algorithm for \earleyFSA.

\subsection{One-Word Lookahead}\label{app:lookahead}

Orthogonally to \cref{app:left-recursion}, we can optionally extend the left child relation to terminal symbols, saying that $a$ is a left child of $A$ if there exists a rule $\edge{A}{a \, \rho}$.  

The resulting extended left-corner relation (in its unweighted version) can be used to construct a side condition on \predone (or \predonemod), so that at position $j$, it does not predict all symbols that are compatible with the left context, but only those that are \emph{also} compatible with the next input terminal.  To be precise, \predone (or \predonemod) should only predict $B$ at position $j$ if $\state{j,k,a}$ and $\leftcorner{B}{a}$ (for some $a$). This is in fact \citet{earley70}'s $k$-word lookahead scheme in the special case $k=1$. 

\subsection{Left-Corner Parsing}

\Citet{nederhof-1993-generalized} and \citet{nederhof-1994-optimal} describe a \defn{left-corner parsing} technique that we could apply to further speed up Earley's algorithm.  This subsumes the one-word lookahead technique of the previous section.  \Citet{eisner-blatz-2007} sketched how the technique could be derived automatically.

Normally, if $B$ is a deeply nested left corner of $C$, then the item $\edge{A}{\mu \,\bigcdot\, C \, \nu}$ will trigger a long chain of \predict actions that culminates in $\fillstate$.  Unfortunately, it may not be possible for this $B$ (or anything predicted from it) to \scan its first terminal symbol, in which case the work has been wasted.  

But recall from \cref{app:left-recursion} that the \predonemod rule effectively summarizes this long chain of predictions using a precomputed weighted item $\leftcorner{C}{B}$.  The left-corner parsing technique simply skips the \predict steps and uses $\leftcorner{C}{B}$ as a side condition to lazily \emph{check after the fact} that the relevant prediction of a $\bigcdot$-initial rule could have been made.  

\predone is removed, so the method never creates dotted productions of the form $\edge{A}{\mu \,\bigcdot\, \nu}$ where $\mu=\varepsilon$---except 
for the start item and the items derived from it using \predtwo.  

In \comptwo, a side condition $\mu \neq \varepsilon$ is added.  
For the special case $\mu=\varepsilon$, a new version of \comptwo is used in which 
\begin{itemize}[nosep,label=--]
\item $i=j$ is required, 
\item the first antecedent $\state{i,i,\edge{A}{\bigcdot\, B\, \nu}}$ is replaced by $\edge{A}{B\, \nu}$ (which ensures that $\state{i,i,\edge{A}{\bigcdot\, B\, \nu}}$ is an item of \earleyDeductionFast),
\item the side conditions $\state{h,i, \edge{D}{\mu' \,\bigcdot\, C \, \nu'}}$ and $\leftcorner{C}{A}$ (which ensures that  \earleyDeductionFast would have \predict{}ed that item).  Note that $\mu'=\varepsilon$ is possible in the case where $D$ is the start symbol $\start$.
\end{itemize}

\noindent The \scan rule is split in exactly the same way into $\mu\neq \varepsilon$ and $\mu=\varepsilon$ variants.

\section{Execution of Weighted \earleyDeductionFast}\label{sec:implementation}\label{app:pseudocode}

\citet{eisner-2023-tacl} presents generic strategies for executing unweighted and weighted deduction systems.  We apply these here to solve the weighted recognition and prefix weight problems, by computing the weights of all items that are provable from given grammar and sentence axioms.

\subsection{Execution via Multi-Pass Algorithms}\label{app:twopass}\label{app:cyclic}

The \earleyDeduction and \earleyDeductionFast deduction systems are nearly acyclic, thanks to our elimination of unary rule cycles and nullary rules from the grammar.  However, cycles in the left-child relation can still create deduction cycles, with $\state{k,k,\edge{A}{\,\bigcdot\,B\,X}}$ and $\state{k,k,\edge{B}{\,\bigcdot\,A\,Y}}$ proving each other via \pred or via \predone and \predtwo.

Weighted deduction can be accomplished for these systems using the generic method of \citet[\S7]{eisner-2023-tacl}.  This will detect the left-child cycles at runtime \cite{tarjan-1972} and solve the weights to convergence within each strongly connected component (SCC).  While solving the SCCs can be expensive in general, it is trivial in our setting since the weights of the items within an SCC do not actually depend on one another: these items serve only as side conditions for one another.  Thus, any iterative method will converge immediately.

Alternatively, the deduction system becomes fully acyclic when we eliminate prediction chains as shown in \cref{app:left-recursion}.  In particular, this modified version of \earleyDeductionFast replaces \predone with \predonemod.\footnote{Recall that eliminating the left-child cycles in advance in this way is needed when one wants to compute weights of the form $\weight{\stategeneric} = (\incinside(\stategeneric),\prefixoutside(\stategeneric))$, in which case the items in an SCC do \emph{not} merely serve as side conditions for one another.  The weighted deduction formalism of \citet{eisner-2023-tacl} is flexible enough to handle cyclic rules that would correctly define these pairwise weights in terms of one another, but solving the SCCs would no longer be fast.}
Using this acyclic deduction system allows a simpler execution strategy: under any acyclic deduction system, a reference-counting strategy \cite{kahn-1962} can be applied to find the proved items and then compute their weights in topologically sorted order \cite[\S6]{eisner-2023-tacl}.

In both cyclic and acyclic cases, the above weighted recognition strategies consume only a constant factor more time and space than their unweighted versions, across all deduction systems and all inputs.\footnote{Excluding the time to solve the SCCs in the cyclic case; but for us, the statement holds even when including that time.}  For \earleyDeductionFast and its acyclic version, this means the runtimes are $\bigo{\nN\abs{\cfg}}$ for a class of ``bounded-state'' grammars, $\bigo{\nN^2\abs{\cfg}}$ for unambiguous grammars, and $\bigo{\nN^3\abs{\cfg}}$ for general grammars (as previewed in the abstract and \cref{sec:intro}).  The space requirements are respectively $\bigo{\nN\abs{\cfg}}$, $\bigo{\nN^2\abs{\cfg}}$, and $\bigo{\nN^2\abs{\cfg}}$.  The same techniques apply to \earleyFSA, replacing $\abs{\cfg}$ with $\abs{\fsa}$.

\subsection{One-Pass Execution via Prioritization}\label{app:lexpriorities}

For the acyclic version of the deduction system (\cref{app:left-recursion}), an alternative strategy is to use a prioritized agenda to visit the items of the acyclic deduction system in some topologically sorted order \cite[\S5]{eisner-2023-tacl}.
This may be faster in practice than the generic reference-counting strategy because it requires only one pass instead of two.  It also remains space-efficient.  On the other hand, it requires a priority queue, which adds a term to the asymptotic runtime (worsening it in some cases such as bounded-state grammars).

We must associate a \defn{priority} $\pi(\stategeneric)$ with each item $\stategeneric$ such that if $\stategenerictwo$ is an antecedent or side condition in some rule that proves $\stategeneric$, then $\pi(\stategenerictwo) < \pi(\stategeneric)$.  Below, we will present a nontrivial prioritization scheme in which the priorities implicitly take the form of lexicographically ordered tuples. 

These priorities can easily be converted to integers in a way that preserves their ordering.  Thus, a bucket queue \cite{dial-1969} or an integer priority queue \cite{thorup-2000} can be used (see \citet[\S5]{eisner-2023-tacl} for details).  The added runtime overhead\footnote{Under the Word RAM model of computation and assuming that priorities fit into a single word.} is $\bigo{M}$ for the bucket queue or $\bigo{M' \log \log M'}$ for the integer priority queue, where $M = \bigo{\nN^2\abs{\cfg}}$ is the number of distinct priority levels in the set of \emph{possible} items, and $M' \leq M$ is the number of distinct priority levels of the \emph{actually} proved items, which depends on the grammar and input sentence.

For \earleyDeductionFast with the modifications of \cref{app:left-recursion}, we assign the minimum priority to 
all of the axioms.  All other items have one of six forms:
\begin{enumerate}[noitemsep]
    \item $\state{j, k, \edge{B}{\rho\ \bigcdot}}$ (antecedent to \compone, \pos)
    \item $\state{j, k, \edge{B}{\star\ \bigcdot}}$ \\(rightmost antecedent to \comptwo)
    \item $\state{j, k, \edge{B}{\mu\,\bigcdot\,\nu}}$ where $\mu \neq \emptystring, \nu \neq \emptystring$ (leftmost antecedent to \predonemod, \scan, \pos)
    \item $\prefixstatek$ (antecedent to nothing)
    \item $\state{k, k, \edge{B}{\bigcdot\ \star}}$ (antecedent to \predtwo)
    \item $\state{k, k, \edge{B}{\bigcdot\ \rho}}$ \\ (leftmost antecedent to \scan, \comptwo) 
\end{enumerate}
The relative priorities of these items are as follows:
\begin{itemize}
\item Items with smaller $k$ are visited sooner (left-to-right processing).  

\item Among items with the same $k$, items with $j < k$ are visited before items with $j=k$.  Thus, the leftmost antecedent of \predonemod precedes its consequent.

\item Among items with the same $k$ and with $j < k$, items with larger $j$ are visited sooner.  Thus, the rightmost antecedent of \comptwo precedes its consequent in the case $i < j$, where a narrower item is used to build a wider one.

\item Among items of the first two forms with the same $k$ and the same $j < k$, $B$ is visited sooner than $A$ if $\rewrites{A}{B}$.  This ensures that the rightmost antecedent of \comptwo precedes its consequent in the case $i=j$ and $\nu=\emptystring$, which completes a unary constituent.

To facilitate this comparison, one may assign integers to the nonterminals according to their height in the unweighted graph whose vertices are $\nonterminals$ and whose edges $\edge{A}{B}$ correspond to the unary productions $\edge{A}{B}$.  (This graph is acyclic once unary cycles have been eliminated by the method of \cref{app:unary}.)

\item Remaining ties are broken in the order of the numbered list above.  This ensures that the antecedents of $\compone$, $\pos$, and $\predtwo$ precede their consequents, and the rightmost antecedent of $\comptwo$ precedes its consequent in the case $i=j$ and $\nu\neq\emptystring$, which starts a non-unary constituent.
\end{itemize}
To understand the flow of information, notice that the 6 specific items in the numbered list above would be visited in the order shown.  

\subsection{Pseudocode for Prioritized Algorithms}\label{app:pseudocode-pqueue}

For concreteness, we now give explicit pseudocode that runs the rules to build all of the items in the correct order.  This may be easier to implement than the above reductions to generic methods. It is also slightly more efficient than 
\cref{app:lexpriorities}, due to exploiting some properties of our particular system.

Furthermore, in this section we handle \earleyDeductionFast as well as its acyclic modification.  When the flag $\prefixtrue$ is set to \textit{true}, we carry out the acyclic version, which replaces \predone with \predonemod and \startrule
(\cref{app:left-recursion}), and also includes \pos (\cref{sec:prefix}) to find prefix weights.

The algorithm pops (dequeues) and processes items in the same order as \cref{app:lexpriorities} (when $\prefixtrue$ is \textit{true}), except that in this version, axioms of the form $\edge{B}{\rho}$ and $\state{k-1, k, a}$ are never pushed (enqueued) or popped but are only looked up in indices.  Similarly, the $\prefixstate$ items (used to find prefix weights) are never pushed or popped but only proved.  Thus, none of these items need priorities.

When an item $\stategenerictwo$ is popped, our pseudocode invokes only deduction rules for which $\stategenerictwo$ might match the \emph{rightmost} antecedent (which could be a side condition), or in the case of \scan or \predonemod, the \emph{leftmost} antecedent.  In all cases, the other antecedents are either axioms or have lower priorities.
While we do not give pseudocode for each rule, invoking a rule on $\stategenerictwo$ always checks first whether $\stategenerictwo$ actually does match the relevant antecedent.  If so, it looks up the possible matches for its other antecedents from among the axioms and the previously proved items.  This may allow the rule to prove consequents, which it adds to the queues and indices as appropriate (see below). 

The main routine is given as \cref{algo:earley-priority}.  A queue iteration such as ``\textbf{for} $k\in\pqueuecol$: \ldots'' iterates over a collection that may change during iteration; it is shorthand for ``\textbf{while} $\pqueuecol \neq \emptyset$: \{ $k=\pqueuecol.\text{pop}()$; \ldots \}.''\looseness=-1

We maintain a dictionary (the \defn{chart}) that maps items to their weights.  Each time an item $\stategeneric$ is proved by some rule, its weight $\weight{\stategeneric}$ is updated accordingly, as explained in \cref{sec:weighted} and \cref{tab:prefix}.  The weight is $\incinside(\stategeneric)$ or $(\incinside(\stategeneric),\prefixoutside(\stategeneric))$ according to whether $\prefixtrue$ is $\textit{false}$ or $\textit{true}$.  

\Cref{algo:earley-priority} writes $\chart(\textit{pattern})$ to denote the set of all provable items (including axioms) that match \textit{pattern}.  This set will have previously been computed and stored in an \defn{index} dedicated to the specific invocation of $\chart(\textit{pattern})$ in the pseudocode.  The index is another dictionary, with the previously bound variables of the pattern serving as the key.  The pseudocode for individual rules also uses indices, to look up antecedents.

When an item $\stategeneric$ is \emph{first} proved by a rule and added to the chart, it is also added to all of the appropriate sets in the indices.  Prioritization ensures that we do not look up a set until it has converged.

Each dictionary may be implemented as a hash table, in which case lookup takes expected $O(1)$ time under the Uniform Hashing Assumption.  An array may also be used for guaranteed $O(1)$ access, although its sparsity may increase the algorithm's asymptotic space requirements.\footnote{Its sparsity need not increase the runtime requirements, however: an \emph{uninitialized} array can be used to simulate an initialized array with constant overhead.  \Citet{higham-schenk-1993} attribute this technique to computer science folklore.}

What changes when $\prefixtrue$ is \textit{false}, other than a few of the rules?  Just one change is needed to the prioritization scheme of \cref{app:lexpriorities}.
The \earleyDeductionFast deduction system is cyclic, as mentioned in \cref{app:cyclic}, so in this case, we cannot enforce $\pi(\stategenerictwo) < \pi(\stategeneric)$ when $\stategenerictwo$ and $\stategeneric$ are an antecedent and consequent of the same rule.  We will only be able to guarantee $\pi(\stategenerictwo) \leq \pi(\stategeneric)$, where the $=$ case arises only for \predone and \predtwo.  To achieve this weaker prioritization, we modify our tiebreaking principle from \cref{app:lexpriorities} (when $\prefixtrue$ is \textit{false}) to say that for a given $k$, all items of the last two forms have \emph{equal} priority and thus may be popped in any order.\footnote{Formerly, all items of form 5 already had equal priority, and so did all items of form 6, but the former priority was strictly lower.  This worked because there were no prediction chains.\looseness=-1}
When a rule proves a consequent that has the same priority as one of its antecedents, it is possible that the consequent had popped previously.  In our case, this happens only for the rule \predone, so crucially, it does not matter if the new proof changes the consequent's weight---this consequent is used only as a side condition (to \predtwo) so its weight is ignored.  However, to avoid duplicate work, we must take care to avoid re-pushing the consequent now that it has been proved again.\footnote{Specifically, this discussion implies that in general, when a consequent may have the same priority as an antecedent, we must check whether it has \emph{ever} been pushed onto the queue, not whether it is currently on the queue.  Luckily, this is easily done by checking whether it is a key in the chart.}  

Rather than place all the items on a single queue that is prioritized lexicographically as in \cref{app:lexpriorities}, we use a collection of priority queues that are combined in the pseudocode to have the same effect.  They are configured and maintained as follows.

\begin{algorithm}[t]
\begin{algorithmic}[1]
\Function{EarleyFast}{$\cfg, \vx, \prefixtrue$}
\State add $\cfg, \vx$ axioms to dictionaries and queues
\IIf{$\prefixtrue$}{$\startrule()$} \Comment{apply \startrule (\cref{app:left-recursion})}
\For{$k \in \pqueuecol$} \Comment{that is: while $\pqueuecol\neq\emptyset$, pop into $k$}
    \For{$j \in \pqueuespanstd$}
        \For{$B \in \pqueueitemstd$} 
            \For{$\stategenerictwo \in \chart(\state{j, k, \edge{B}{\rho\ \bigcdot}})$} \Comment{form 1}
                \State $\compone(\stategenerictwo)$; $\pos(\stategenerictwo)$
            \EndFor
            \For{$\stategenerictwo \in \chart(\state{j, k, \edge{B}{\star\ \bigcdot}})$} \Comment{form 2}
                \State $\comptwo(\stategenerictwo)$
            \EndFor
        \EndFor
        \For{$\stategenerictwo \in \chart(\state{j, k, \edge{B}{\mu\bigcdot\nu}}, \mu \!\!\neq\!\! \emptystring \!\!\neq\!\! \nu)$}
            \State $\scan(\stategenerictwo)$; $\pos(\stategenerictwo)$ \Comment{form 3}
            \IIfElse{$\prefixtrue$}{$\predonemod(\stategenerictwo)$}{$\predone(\stategenerictwo)$} 
        \EndFor
    \EndFor
    \If{$\prefixtrue$ \textbf{and} $k>0$} \Comment{form 4}
       \State now $w(\prefixstatek) = \mbox{}$prefix weight of $\vx_{0:k}$
    \EndIf
    \If{$\prefixtrue$}
       \For{$\stategenerictwo \in \chart(\state{k, k, \edge{B}{\bigcdot\ \star}})$} \Comment{form 5}
           \State $\predtwo(\stategenerictwo)$
       \EndFor
       \For{$\stategenerictwo \in \chart(\state{k, k, \edge{B}{\bigcdot\ \rho}})$} \Comment{form 6}
           \State $\scan(\stategenerictwo)$
       \EndFor
    \Else
        \For{$\stategenerictwo \in \predicted{k}$} \Comment{forms 5 and 6}
           \LinesComment{prediction may push new items onto $\predicted{k}$}           
           \State $\predone(\stategenerictwo)$; $\predtwo(\stategenerictwo)$; $\scan(\stategenerictwo)$ 
        \EndFor
    \EndIf
\EndFor
\For{$\stategenerictwo \in \chart(\state{0,\abs{\vx},\edge{S}{\,\star\,\bigcdot}})$}
   \State \Return{$\weight{\stategenerictwo}$} \Comment{weight of goal item}
\EndFor
\State \Return{$\zero$} \Comment{goal item has not been proved}
\EndFunction
\end{algorithmic}
\caption{\earleyDeductionFast with priority queues}
\label{algo:earley-priority}
\end{algorithm}

\begin{itemize}
\item $\pqueuecol$ is a priority queue of distinct positions $k \in \{0,\dots, N\}$, which pop in increasing order.  $k$ is added to it upon proving an item of the form $\state{\cdot,k,\cdot}$.  Initially $\pqueuecol=\{0\}$ due to the start axiom $\state{0,0,\edge{S}{\,\bigcdot\, \star}}$.
\item For each $k \in \pqueuecol$, $\pqueuespanstd$ is a priority queue of distinct positions $j \in \{0,\dots,k\}$, which pop in decreasing order except that $k$ itself pops last.  $j$ is added to it upon proving an item of the form $\state{j,k,\cdot}$.  Initially $\pqueuespan{0} = \{0\}$ due to the start axiom.
\item For each $j \in \pqueuespanstd$ with $j < k$,  $\pqueueitemstd$ is a priority queue of distinct nonterminals $B \in \nonterminals$, which pop in the height order described in \cref{app:lexpriorities} above.  $B$ is added to it upon proving an item of the form $\state{j, k, \edge{B}{\rho\ \bigcdot}}$.
\item If $\prefixtrue$ is \textit{false}, then for each $k \in \pqueuecol$, $\predicted{k}$ is a queue of all proved items of the form $\state{k, k, \edge{B}{\bigcdot\ \star}}$ or $\state{k, k, \edge{B}{\bigcdot\ \rho}}$.  These items have equal priority so may pop in any order (e.g., LIFO).  Initially $\predicted{0}$ contains just the start axiom.
\end{itemize}

Transitive consequents added later to a queue always have $\geq$ priority than their antecedents that have already popped, so the minimum priority of the queue increases monotonically over time.  This monotone property is what makes bucket queues viable in our setting \cite[see][\S5]{eisner-2023-tacl}.  In general, our priority queues are best implemented as bucket queues if they are dense and binary heaps or integer priority queues if they are sparse.

\section{Binarized \earleyFSA}\label{app:fsa-binarized}

\Cref{tab:fsa-binarized} gives a version of $\earleyFSA$ in which the ternary deduction rules $\scan$, $\compone$ and $\comptwo$ have been binarized using the fold transform, as promised in \cref{sec:fsa}.
\begin{itemize}
\item The $\scanone$ and $\scantwo$ rules, which replace $\scan$, introduce and consume new intermediate items of the form $\state{i, j, \dfaEdge{q}{a}{\star}}$.  The $\scanone$ rule sums over possible start positions $j$ for word $a$.  This is only advantageous in the case of lattice parsing (see \cref{fn:lattice}), since for string parsing, the only possible choice of $j$ is $k-1$.
\item In a similar vein, $\comptwoa$ and $\comptwob$ introduce and consume new intermediate items $\state{i, j, \dfaEdge{\star}{A}{q}}$.  The $\comptwoa$ rule aggregates different items from $i$ to $j$ that are looking for a $B$ constituent to their immediate right, summing over their possible current states $q$.
\item Similarly, $\componea$ introduces new intermediate items that sum over possible final states $q'$.
\item We did not bother to binarize the ternary rule $\filter$, as there is no binarization that provides an asymptotic speed-up. 
\end{itemize}

There are different ways to binarize inference rules, and in \cref{tab:fsa-binarized} we have chosen to binarize $\scan$ and $\comptwo$ in complementary ways.  Our binarization of $\scan$ is optimized for the common case of a dense WFSA and a sparse sentence, where state $q$ allows many terminal symbols $a$ but the input allows only one (as in string parsing) or a few.  $\scanone$ finds just the symbols $a$ allowed by the input and $\scantwo$ looks up only those out-arcs from $q$.  Conversely, our binarization of $\comptwo$ is optimized for the case of a sparse WFSA and a dense parse table: $\comptwoa$ finds the small number of incomplete constituents over $\state{i,j}$ that are looking for a $B$, and $\comptwob$ looks those up when it finds a complete $B$ constituent, just like \earleyDeductionFast.

It is possible to change each of these binarizations.  In particular, binarizing $\scan$ by first combining
$\state{i,j,q}$ with $\dfaEdge{q}{a}{q'}$ (analogously to $\comptwoa$) would be useful when parsing a large or infinite lattice---such as the trie implicit in a neural language model---with a constrained grammar \citep{semanticmachines-2021-emnlp, fang-2022-whole-truth}.

\renewcommand{\arraystretch}{2.4}

\begin{table*}[t]
\centering
\small
\begin{tabularx}{0.98\linewidth}{l l l}
    \toprule
    \bf  Domains & \multicolumn{2}{l}{$i,j,k\in\{0,\dots,\nN\} \quad A\in\nonterminals \quad a\in\terminals \quad q,q'\in\dfaStates$} \\
    \bf Items & \multicolumn{2}{l}{$\state{i, j, q} \quad \state{i, j, q?} \quad \state{i, j, a} \quad \state{i, j, \edge{A}{\,\bigcdot\, \star}} \quad \state{i, j, \edge{A}{\star \,\bigcdot\,}}
    \quad \state{i, j, \dfaEdge{q}{a}{\star}}
    \quad \state{i, j, \dfaEdge{\star}{A}{q}}
    $} 
    \\ 
        &
    \multicolumn{2}{l}{$\underbrace{q\in\dfaStart \quad q'\in\dfaFinish \quad \dfaEdge{q}{a}{q'} \quad \dfaEdge{q}{A}{q'} \quad \dfaEdge{q}{A}{\star} \quad \dfaEdge{q}{*\hat{A}}{\star} \quad \dfaEdge{q}{\dfaEnd{B}}{\dfaFinish}}_{\text{WFSA items}}$}
    \\    
    \bf Axioms &  \multicolumn{2}{l}{$\state{k-1, k, \x{k}}, \forall k \!\in\! \{1,\dots,\nN \} \quad\quad \state{0, 0, \edge{S}{\bigcdot\,\star}}$\quad\text{WFSA items derived from the WFSA grammar (see \cref{sec:fsa})}}  \\ 
    \bf Goals & $\state{0, \nN, \edge{S}{\star \,\bigcdot\,}}$

    \\ \midrule
    & 
        \AxiomC{}
        \RightLabel{\makecell[l]{$\state{i, j, q}$ \, $\dfaEdge{q}{B}{\star}$}}
        \LeftLabel{\smaller \predone:}
        \UnaryInfC{$\state{j, j, \edge{B}{\,\bigcdot\, \star}}$}
        \DisplayProof
    & 
        \def\defaultHypSeparation{\hspace{0pt}}
        \AxiomC{$\dfaEdge{q}{\dfaEnd{B}}{q'}$}
        \AxiomC{$q'\in\dfaFinish$}
        \LeftLabel{\smaller \componea:}
        \BinaryInfC{$\dfaEdge{q}{\dfaEnd{B}}{\dfaFinish}$}
        \DisplayProof \\        
    & 

        \AxiomC{$q\in\dfaStart$}
        \LeftLabel{\smaller \predtwo:}
        \UnaryInfC{$\state{j, j, q?}$}
        \DisplayProof
    & 
        \def\defaultHypSeparation{\hspace{0pt}}
        \AxiomC{$\state{j, k, q}$}
        \AxiomC{$\dfaEdge{q}{\dfaEnd{B}}{\dfaFinish}$}
        \LeftLabel{\smaller \componeb:}
        \BinaryInfC{$\state{j, k, \edge{B}{ \star \,\bigcdot\,}}$}
        \DisplayProof

    \\
    \textbf{Rules}
    &
        \hspace{-7pt}
        \def\defaultHypSeparation{\hspace{0pt}}
        \AxiomC{$\state{i, j, q}$}
        \AxiomC{$\state{j, k, a}$}
        \LeftLabel{\smaller \scanone:}
        \BinaryInfC{$\state{i, k, \dfaEdge{q}{a}{\star}}$}
        \DisplayProof

    &
        \def\defaultHypSeparation{\hspace{0pt}}
        \AxiomC{$\state{i, j, q}$}

        \AxiomC{$\dfaEdge{q}{B}{q'}$}
        \LeftLabel{\smaller \comptwoa:}
        \BinaryInfC{$\state{i, j, \dfaEdge{\star}{B}{q'}}$}

        \DisplayProof

    \\
    & 
        \hspace{-7pt}
        \def\defaultHypSeparation{\hspace{0pt}}

        \AxiomC{$\state{i, k, \dfaEdge{q}{a}{\star}}$}
        \AxiomC{$\dfaEdge{q}{a}{q'}$}
        \LeftLabel{\smaller \scantwo:}
        \BinaryInfC{$\state{i, k, q'?}$}

        \DisplayProof
    &
        \def\defaultHypSeparation{\hspace{0pt}}

        \AxiomC{$\state{i, j, \dfaEdge{\star}{B}{q'}}$}
        \AxiomC{$\state{j, k, \edge{B}{ \star \,\bigcdot\,}}$}
        \LeftLabel{\smaller \comptwob:}
        \BinaryInfC{$\state{i, k, q'?}$}

        \DisplayProof
    \\
    &

        \hspace{-7pt}
        \def\defaultHypSeparation{\hspace{0pt}}
        \AxiomC{$\state{i, j, q}$}
        \AxiomC{$\dfaEdge{q}{\emptystring}{q'}$}
        \LeftLabel{\smaller \eps:}
        \BinaryInfC{$\state{i, j, q'?}$}
        \DisplayProof

    &

        \def\defaultHypSeparation{\hspace{0pt}}
        \AxiomC{$\state{j, k, q?}$}
        \RightLabel{\makecell[l]{$\state{j, j, \edge{A}{\,\bigcdot\, \star}}$ \, $\dfaEdge{q}{*\hat{A}}{\star}$}}
        \LeftLabel{\smaller \filter:}
        \UnaryInfC{$\state{j, k, q}$}
        \DisplayProof

    \\ \bottomrule
\end{tabularx}
\caption{Improvement of \earleyFSA (\cref{tab:fsa}) in which $\scan$, $\compone$ and $\comptwo$ have been binarized using a fold transform. Since $\componea$ does not depend on the input it can in practice be run during preprocessing, just like the rules that derive other WFSA items such as $\dfaEdge{q}{*\hat{A}}{\star}$. 
See the main text (\cref{app:fsa-binarized}) for a discussion of alternative binarization schemes.}
\label{tab:fsa-binarized}
\end{table*}

\section{Handling Nullary and Unary Productions in an FSA}\label{app:fsa-null}

As for \earleyDeductionFast, \earleyFSA (\cref{sec:fsa}) requires elimination of nullary productions. We can handle nullary productions by directly adapting
the construction of \cref{app:null} to the WFSA case.  Indeed, the WFSA version is simpler to express. For each arc $\dfaEdge{q}{B}{q'}$ such that $B\in\nonterminals$ and $\e{B}\neq\zero$, we replace the $B$ label of that arc with $\noneps{B}$ (preserving the arc's weight), and add a new arc $\dfaEdge{q}{\emptystring}{q'}$ of weight $\e{B}$.
We then define a new WFSA $\fsa' = (\fsa \cap \neg \fsa_{\mathrm{bad}}) \cup \fsa_{\mathrm{good}}$, where $\fsa_{\mathrm{bad}}$ is an unweighted FSA that accepts exactly those strings of the form $\hat{A}$ (i.e., nullary productions), $\neg$ takes the unweighted complement, and $\fsa_{\mathrm{good}}$ is a WFSA that accepts exactly strings of the form $\hat{S}$ (with weight $\e{S}$) and $\noneps{S}\hat{S}$ (with weight $\one$).  As this construction introduces new $\emptystring$ arcs, it should precede the elimination of $\emptystring$-cycles.

Notice that in the example of \cref{app:null} where a production $\edge{A}{\rho}$ was replaced with up to $2^k-1$ variants, the WFSA construction efficiently shares structure among these variants.  It adds at most $k$ edges at the first step and at most doubles the total number of states through intersection with $\neg \fsa_{\mathrm{bad}}$.\saveforcameraready{}

Similarly, we can handle unary productions by directly adapting the construction of \cref{app:unary} to the WFSA case.
We first extract all weighted unary rules by intersecting $\fsa$ with the unweighted language $\{B\hat{A}: A,B \in \nonterminals\}$ (and determinizing the result so as to combine duplicate rules).  Exactly as in \cref{app:unary}, we construct the unary rule graph and compute its SCCs along with weights $\weight{\rewrites{A}{B}}$ for all $A,B$ in the same SCC.
We modify the WFSA by underlining all hatted nonterminals $\hat{A}$ and overlining all nonterminals $B$.
Finally, we define our new WFSA grammar $(\fsa \cap \neg \fsa_{\mathrm{bad}}) \cup \fsa_{\mathrm{good}}$.  Here $\fsa_{\mathrm{bad}}$ is an unweighted FSA that accepts exactly those strings of the form $\overline{B}\hat{\underline{A}}$ and $\fsa_{\mathrm{good}}$ is a WFSA that accepts exactly strings of the form $\underline{B}\hat{\overline{A}}$ such that $A,B$ are in the same SCC, with weight $\weight{\rewrites{A}{B}}$.

Following each construction, nonterminal names can again be simplified as in \cref{app:unary,app:null}.

Finally, \cref{sec:fsa} mentioned that we must eliminate $\emptystring$-cycles from the FSA.  The algorithm for doing so \citep{mohri02} is fundamentally the same as our method for eliminating unary rule cycles from a CFG (\cref{app:unary}), but now it operates on the graph whose edges are $\emptystring$-arcs of the FSA, rather than the graph whose edges are unary rules of the CFG.  

\saveforcameraready{}

\section{Non-Commutative Semirings}\label{app:noncomm}

We finally consider the case of non-commutative weight semirings, where the order of multiplication becomes significant.

In this case, in the product \labelcref{eq:treeprod} that defines the weight of a derivation tree $T$, the productions should be multiplied in the order of a pre-order traversal of $T$.

In \cref{sec:deduction}, when we recursively defined the weight $w(d_V)$ of a proof, we took a product over the above-the-bar antecedents of a proof rule.  These should be multiplied in the same left-to-right order that is shown in the rule.  Our deduction rules are carefully written so that under these conventions, the resulting proof weight matches the weight \labelcref{eq:treeprod} of the corresponding CFG derivation.  

For the same reason, the same left-to-right order should be used in \cref{sec:deduction} when computing the inside probability $\incinside(V)$ of an item.

Eliminating nullary productions from a weighted CFG (\cref{app:null}) is not in general possible in non-commutative semirings.  However, if the grammar has no nullary productions or is converted to an FSA before eliminating nullary productions (\cref{app:fsa-null}), then weighted parsing may remain possible.

What goes wrong?  The construction in \cref{app:null} unfortunately reorders the weights in the product \labelcref{eq:treeprod}.  Specifically, in the production $\edge{A}{\mu\, B\, \nu}$, the product should include the weight $\e{B}$ \emph{after} the weights in the $\mu$ subtrees, but our construction made it part of the weight of the modified production $\edge{A}{\mu\, \nu}$ and thus moved it \emph{before} the $\mu$ subtrees.  This is incorrect when $\mu\neq\emptystring$ and $\otimes$ is non-commutative.  

The way to rescue the method is to switch to using WFSA grammars (\cref{sec:fsa}).  The WFSA grammar breaks each rule up into multiple arcs, whose weights variously fall before, between, and after the weights of its children.  When defining the weight of a derivation under the WFSA grammar, we do not simply use a pre-order traveral as in \cref{eq:treeprod}.  The definition is easiest to convey informally through an example.  Suppose a derivation tree for $\rewrites{A}{\vx}$ uses a WFSA path at the root that accepts $BC\dfaEnd{A}$ with weight $w$. Recursively let $w_B$ and $w_C$ be the weights of the child subderivations, rooted at $B$ and $C$.  Then the overall weight of the derivation of $A$ will not be $w \otimes w_B \otimes w_C$ (prefix order), but rather $w_1 \otimes w_B \otimes w_2 \otimes w_C \otimes w_3$.  Here we have factored the path weight $w$ into $w_1 \otimes w_2 \otimes w_3$, which are respectively the weights of the subpath up through $B$ (including the initial-state weight), the subpath from there up through $C$, and the subpath from there to the end (including the final-state weight).

When converting a CFG to an equivalent WFSA grammar (\cref{fn:encodecfg}), the rule weight always goes at the \emph{start} of the rule so that the weights are unchanged.  However, the nullary elimination procedure for the WFSA (\cref{app:fsa-null}) is able to replace unweighted nonterminals in the \emph{middle} of a production with weighted $\varepsilon$-arcs.  This is the source of its extra power, as well as its greater simplicity compared to \cref{app:null}.

It really is not possible to fully eliminate nulls within the simpler weighted CFG formalism.  
Consider an unambiguous weighted CFG whose productions are $\edge{S}{a\,S\,A}, \edge{S}{b\,S\,B}, \edge{S}{c}, \edge{A}{\emptystring}, \edge{B}{\emptystring}$, with respective weights $w_a, w_b, w_c, w_A, w_B$.  Then a string $\vx = abbc$ will have $Z_\vx$ given by the mirrored product $w_a \otimes w_b \otimes w_b \otimes w_c \otimes w_B \otimes w_B \otimes w_A$.  Within our weighted CFG formalism, there is no way to include the final weights $w_B \otimes w_B \otimes w_A$ if we are not allowed to have null constituents in those positions.  

Even with WFSAs, there is still a problem---in the non-commutative case, we cannot eliminate unary rule cycles (\cref{app:fsa-null}).  If we had built a binary $A$ constituent with weight $w$, then a unary CFG rule $\edge{A}{A}$ with weight $w_1$ required us to compute the total weight of all derivations of $A$, by taking a summation of the form $w \oplus (w_1 \otimes w) \oplus (w_1 \otimes w_1 \otimes w) \oplus \cdots$.  
This factors as $(\one \oplus w_1 \oplus (w_1 \otimes w_1) \oplus \cdots) \otimes w$, and unary rule cycle elimination served to precompute the parenthesized sum, which was denoted as $\weight{\rewrites{A}{A}}$, and record it as the weight of a new rule $\edge{\bar{A}}{\underline{A}}$.   However, in the non-commutative case, the WFSA path corresponding to $\edge{A}{A}$ might start with $w_1$ and end with $w_2$.  In that case, the necessary summation has the form $w \oplus (w_1 \otimes w \otimes w_2) \oplus (w_1 \otimes w_1 \otimes w \otimes w_2 \otimes w_2) \oplus \cdots$.  Unfortunately this cannot be factored as before, so we cannot precompute the infinite sums as before.\footnote{A related problem would appear in trying to generalize the left-corner rewrite weights in \cref{app:left-recursion} to the non-commutative case.}  The construction in \cref{app:fsa-null} assumed that we could extract weighted unary rules from the WFSA, with a single consolidated weight at the start of each rule---but consolidating the weight in that way required commutativity. 

\section{Runtime Experiment Results}\label{app:experiment}

\begin{figure}[t]
\centering
\includegraphics[width=\linewidth]{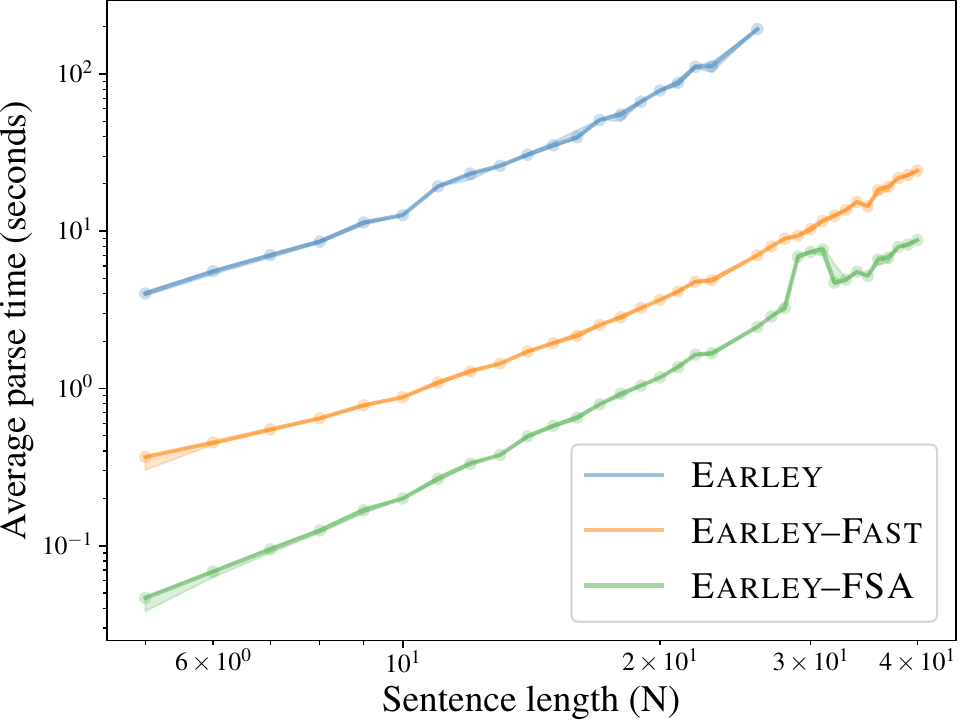}
\newline
\includegraphics[width=\linewidth]{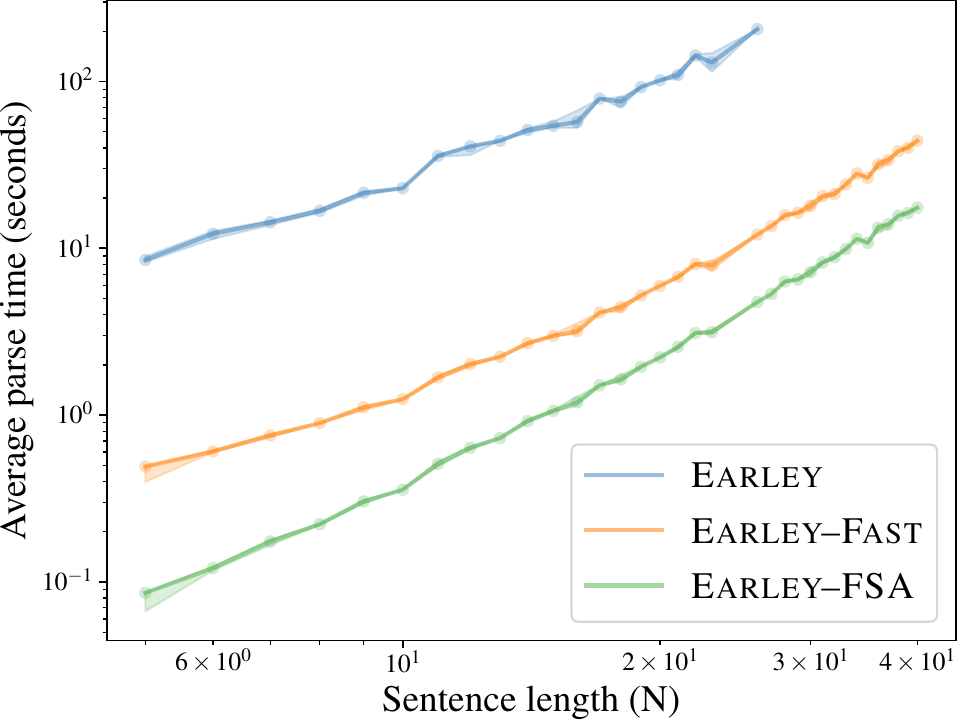}
\caption{Average parse time per sentence for $100$ randomly selected sentences of $5$--$40$ words on the M2 grammar (left) and PM2 grammar (right). As all these algorithms are worst-case cubic in $N$, each curve on these log-log plots is bounded above by a line of slope 3, but the lower lines have better grammar constants. The experiment was conducted using a Cython implementation on an Intel(R) Core(TM) i7-7500U processor with 16GB RAM.}
\label{fig:experiment}
\end{figure}

More details on the experiments of \cref{sec:exp} appear in \cref{fig:experiment}.

\saveforcameraready{
    }

\end{document}